\documentclass{article} 
\usepackage{iclr2023_conference,times}

\usepackage{amsmath,amsfonts,bm}

\def\eqref#1{equation~\ref{#1}}

\def\1{\bm{1}}

\DeclareMathAlphabet{\mathsfit}{\encodingdefault}{\sfdefault}{m}{sl}
\SetMathAlphabet{\mathsfit}{bold}{\encodingdefault}{\sfdefault}{bx}{n}

\newcommand{\R}{\mathbb{R}}

\usepackage{hyperref}
\usepackage{url}
\usepackage{amsmath}
\usepackage{graphicx}
\usepackage{booktabs} 

\usepackage{floatrow}
\DeclareFloatVCode{beforefloat}{\vspace{-0pt}}
\DeclareFloatVCode{afterfloat}{\vspace{-10pt}}
\title{Understanding Neural Coding on Latent Manifolds by Sharing Features and Dividing Ensembles}

\author{Martin Bjerke\textsuperscript{1}, Lukas Schott\textsuperscript{2}, Kristopher T. Jensen\textsuperscript{3},\\
\textbf{Claudia Battistin\textsuperscript{1}, David A. Klindt\textsuperscript{*, 1}, Benjamin A. Dunn\textsuperscript{*, 1}} \\
\textsuperscript{1}Norwegian University of Science and Technology, \textsuperscript{2}Bosch Center for Artificial Intelligence, \\
\textsuperscript{3}University of Cambridge, \textsuperscript{*}Equal contribution\\
\texttt{\{martin.bjerke,benjamin.dunn\}@ntnu.no, klindt.david@gmail.com}
}

\iclrfinalcopy 
\begin{document}

\maketitle

\begin{abstract}
Systems neuroscience relies on two complementary views of neural data, characterized by single neuron tuning curves and analysis of population activity. These two perspectives combine elegantly in neural latent variable models that constrain the relationship between latent variables and neural activity, modeled by simple tuning curve functions. This has recently been demonstrated using Gaussian processes, with applications to realistic and topologically relevant latent manifolds. Those and previous models, however, missed crucial shared coding properties~of neural populations. We propose \emph{feature sharing} across neural tuning curves~which significantly improves performance and helps optimization. We also propose a solution to the \emph{ensemble detection} problem, where different groups of neurons, i.e., ensembles, can be modulated by different latent manifolds. Achieved through a soft clustering of neurons during~training, this allows for the separation of mixed neural populations in an unsupervised~manner. These innovations lead to more interpretable models of neural population activity that train well and perform better even on mixtures of complex latent manifolds. Finally, we apply our method on a~recently published grid cell dataset, and recover distinct ensembles, infer~toroidal latents and predict neural tuning curves in a single integrated modeling~framework.
\end{abstract}



\section{Introduction}

Neural population activity can appear high-dimensional \citep{stringer2019high}, yet much recent work has reported that neural populations in higher brain areas are often confined to low dimensional subspaces \citep{yu2008gaussian,harvey2012choice,mante2013context,stokes2013dynamic,shenoy2013cortical,kaufman2014cortical,sadtler2014neural,gallego2017neural,elsayed2017structure,gao2017theory}.
The bread and butter of classic systems neuroscience is linking neural activity to experimentally controlled or observable covariates such as orientation \citep{hubel1979brain}, pitch \citep{lewicki2002efficient}, movement \citep{churchland2012neural,kao2015single}, posture \citep{mimica2018efficient} and orientation in space \citep{taube1990head}. 
These two parallel streams of neuroscientific research might at first seem to be at odds with each other \citep{kriegeskorte2021neural}; tuning studies of individual neurons give a very different picture of neural coding than distributed representations over high-dimensional neural populations.
However, they combine elegantly in the form of (neural) latent variable models \citep[LVMs, see][]{lawrence2003gaussian,yu2008gaussian,pandarinath2018inferring}.

In their basic form, neural LVMs find the low-dimensional structure of neural population activity, for instance, when a large network of neurons is coding mostly along few linear subspaces \citep{mante2013context,gao2017theory}. 
One advantage is that these models can help us discover latent variables which may not be tracked as classical covariates in systems neuroscience.
However, when the mapping from latent variables to predicted spike rate (decoding) is fully unconstrained, e.g., by using a multi-layer neural network, we lose the simple biological interpretation of tuning curves.

\begin{figure}[t]
  \centering
  \includegraphics[width=.97\textwidth]{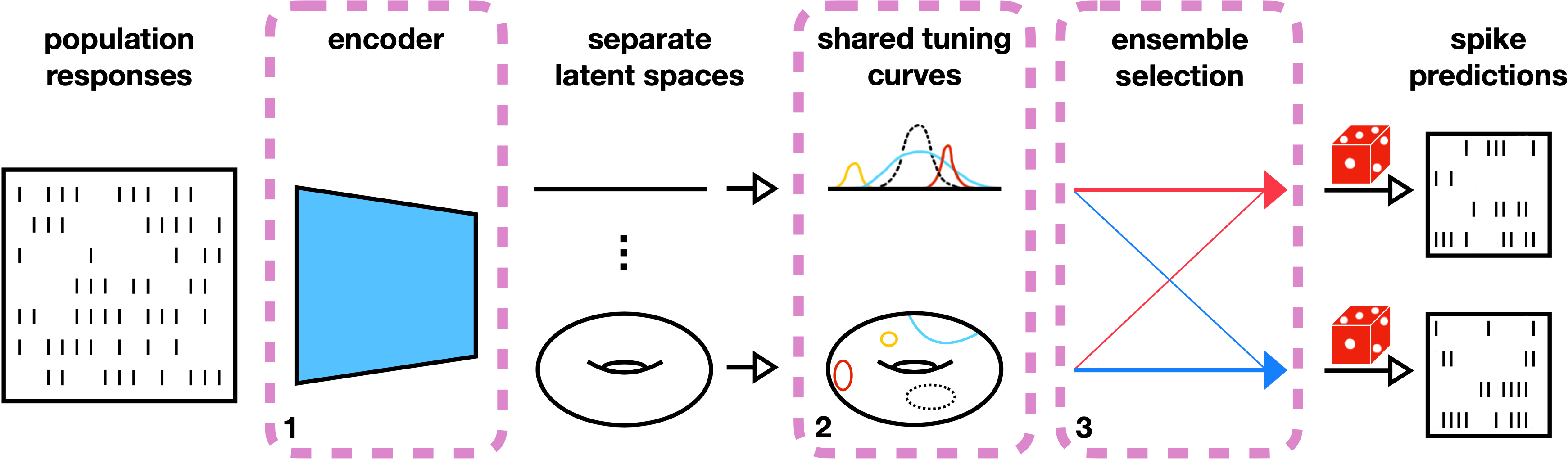}
  \caption{\textbf{Model outline}. Our main contributions are outlined in purple. The input is a matrix~of neural spiking activity. The encoder ($\mathbf{1}$) is a multilayered neural network. The latent spaces are separate with, potentially, different topologies (e.g. $\mathbb{R}^1$ and $\mathbb{T}^2$). The decoder is a parametric tuning curve model with feature sharing ($\mathbf{2}$). The ensemble detection is a weighted (ideally one-hot) selection of latent spaces for each neuron ($\mathbf{3}$). For decoding of the activity, we assume Poisson spiking.\vspace{-0pt}}
  \label{Model_schematic}
\end{figure}

In an effort to maintain a biologically interpretable relationship between the latent variables and the neural activity, recent work has proposed more constrained decoders approximating simple tuning curves.
These tuning curves have been parameterized as Gaussian processes in the framework of Gaussian process latent variable models \citep[GPLVM,][]{wu2017gaussian, wu2018learning}.
Through the tuning curve approach, we limit ourselves to biologically plausible solutions that reveal the actual algorithmic structure of the neural system.
Thus, LVMs with simple tuning curve decoders bring together the view on neural populations as distributed representations of low-dimensional latent variables, along with the biologically meaningful perspective on individual neural tuning properties.

Some neural populations exhibit topologically interesting latent manifolds \citep{singh2008topological,peyrache2015internally,gardner2022toroidal}.
For instance, grid cells represent navigational space in toroidal coordinates of spatially repeating two dimensional hexagonal grids \citep{hafting2005microstructure}.
They appear in different ensembles, commonly referred to as modules, each coding for space at different resolutions \citep{fyhn2007hippocampal,stensola2012entorhinal}.
Thus, the complete population of grid cells is best described as a collection of ensembles of neurons, where neurons in each ensemble have tuning curves of specific shapes on their respective toroidal latent representations of space \citep{curto2017can}.
By contrast, a two-dimensional Euclidean representation might also account for the~complete population of grid cells, but would also completely obscure their efficient and theoretically interesting coding scheme for representing space \citep{solstad2006grid,sreenivasan2011grid,mathis2012optimal,wei2015principle,klukas2020efficient}.
A driving motivation behind this work~was to model this beautiful neural structure with an LVM that separates the algorithmic and biological parts, while uniting shared tuning properties to be more accurate and trainable than previous approaches.

We propose to train neural LVMs that not only have simple tuning curve decoders, but are also fully differentiable.
Thus, we use a flexible encoder, i.e., a neural network as in variational autoencoders \citep{kingma2013auto, rezende2014stochastic}, and a simple tuning curve based decoder, akin to GPLVM.
The encoder can readily be made convolutional to allow for better latent estimation from adjacent time points.
Additionally, we implement a feature basis for the tuning curve shapes in the decoder which can be shared across neurons.
We demonstrate that the neural feature sharing, along with the variational end-to-end training, vastly improves both the training stability as well as the final performance of neural LVMs.
Moreover, we propose hybrid inference at test-time and show that this, again, brings a considerable improvement in performance.
Finally, we integrate the problem of separating distinct ensembles of neurons into our approach --- a crucial task for the discovery of different biological structures and the precise mathematical understanding of their topological tuning properties.
An illustration of our approach is provided in Fig. \ref{Model_schematic}.

To summarize, our full model performs the task of finding latent variables, separating distinct ensembles of neurons and fitting the prototypical tuning curves on each ensemble's latent space in a single efficient framework.
In the following, we therefore refer to our model as \textbf{f}eature sharing \textbf{a}nd \textbf{e}nsemble detection \textbf{L}atent \textbf{V}ariable \textbf{M}odel or \texttt{faeLVM}.

\section{Background}

Let $\lambda_i$ be the instantaneous firing rate of a neuron $i$.
To relate this to the spiking activity $x_i$, we assume a Poisson noise model $x_i \sim \mathcal{P}(\lambda_i)$.
We define latent variables $z:= \{z_1, \hdots, z_k\}$ (in distinct spaces and possibly with different topologies) that change over time and to which neurons are tuned.
More precisely, we suppose that there exists a deterministic function $f_i$, which for every neuron $i$ relates the latent variables to the firing rate, i.e., $\lambda_i = f_i(z)$.
One additional assumption is that each $f_i$ is only \emph{sensitive} to a subset $J_i$ of latent variables and \emph{invariant} to all others.
That means, for two values $z, \hat{z}$ with $z_j = \hat{z}_j \, \forall j \in J_i$ and $z_j \neq \hat{z}_j \, \forall j \notin J_i$ we have $f_i(z) = f_i(\hat{z})$.

As a concrete example, consider a mixed population of head direction neurons $I_1$ and a grid cell module $I_2$.
The former are tuned to a circular latent variable $z_1 \in \mathbb{S}^1$ \citep{peyrache2015internally,rybakken2019decoding,rubin2019revealing,chaudhuri2019intrinsic}, the latter to a toroidal latent~variable $z_2 \in \mathbb{T}^2$ \citep{mcnaughton2006path,fuhs2006spin,gardner2022toroidal}.
That means, for every neuron in the head direction \emph{ensemble} we have $J_i = \{1\}, \, \forall i \in I_1$; whereas for every~neuron in the grid cell \emph{ensemble} we have $J_i = \{2\}, \, \forall i \in I_2$.
Conceivably, there exist shared latent~variables, e.g. attentional state, running speed or pupil dilation, to which multiple or even all neurons~are~tuned.

Within each ensemble, the tuning functions $f_i, \, i \in I_j$ are likely to have similar structure \citep{albright1984direction,taube1990head,hafting2005microstructure}.
This is the concept of \emph{feature sharing} \citep{klindt2017neural}, which in visual systems neuroscience has proven to be a useful assumption \citep{batty2017multilayer,sinz2018stimulus,ecker2018rotation,walker2018inception,cadena2019well,ustyuzhaninov2019rotation,cotton2020factorized,zhuang2021unsupervised,burg2021learning,bashiri2021flow,safarani2021towards,franke2021behavioral,lurz2021generalization,nayebi2021goal,seeliger2021end,goldin2021context,ustyuzhaninov2022digital}.
In vision, this can be motivated by the fact that visual (especially early retinal) neurons form mosaics, that tile the input space and compute the same response function (i.e., tuning curve) across space \citep{wassle1978mosaic}.
Precisely, let $f_{\mu}$ denote the response function of a visual neuron with receptive field center $\mu \in \mathbb{R}^2$, let $T: \mathbb{R}^2 \rightarrow \mathbb{R}^2$ be a spatial translation, and let $z: \mathbb{R}^2 \rightarrow \mathbb{R}^1$ be an image (with slight abuse of notation, $z$ is, here, a function that assigns a grey scale value to every position in two-dimensional space, i.e., an image).
Let us think of $f_i = f_{\mu}$ and $f_{i'} = f_{T(\mu)}$ as two neurons of the same type (i.e., same tuning curve shape) but with their receptive field centers at different locations $\mu$ and $T(\mu)$.
Then we have $f_i(z) = f_{i'}(z \circ T)$.
Thus, for visual neurons we assume \emph{translational equivariance}.

\begin{figure}[t]
\floatbox[{\capbeside\thisfloatsetup{capbesideposition={right,center},capbesidewidth=0.36\textwidth}}]{figure}[\FBwidth]
{\caption{\textbf{The feature sharing assumption}. \textbf{A}, Tuning curves for $26$ head direction tuned neurons. \textbf{B}, The same $26$ tuning curves, shifted and scaled to visualize structural similarity, along with a standard Gaussian shape (dotted line).}\label{Why_feature_sharing_is_a_good_idea}}
{\includegraphics[width=0.60\textwidth]{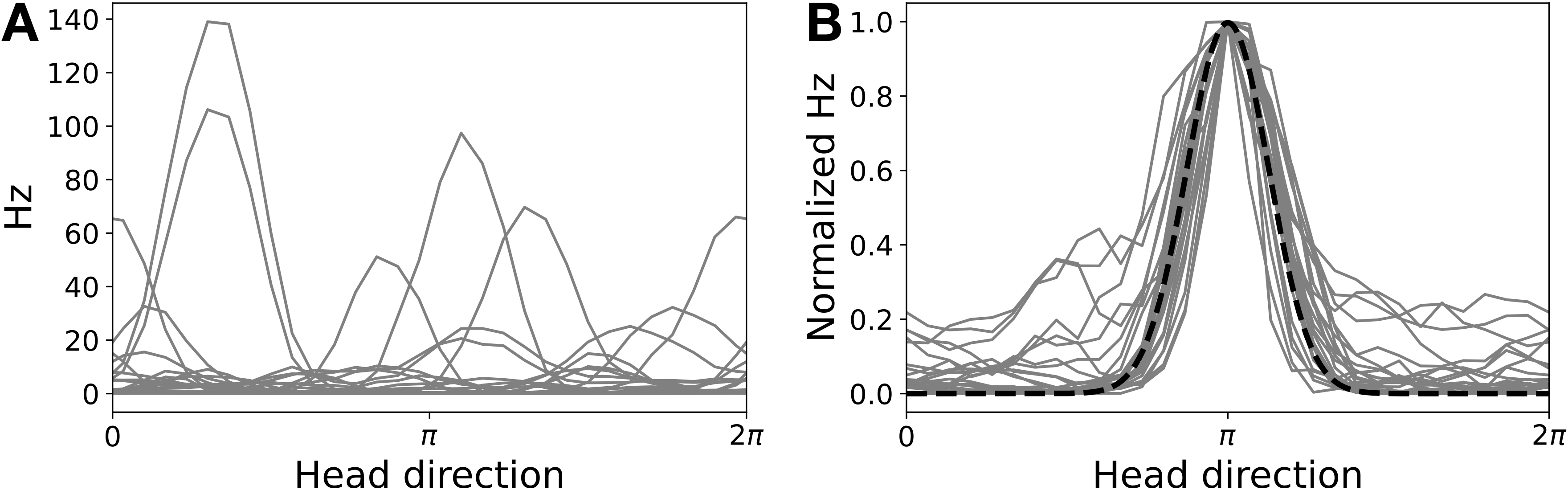}}
\vspace{10pt}
\end{figure}

Consequently, we should be able to marginalize over spatial translations to learn the shared structure among all $f_i$ as demonstrated in \citet{klindt2017neural}.
The same intuition holds for populations of auditory neurons that are equivariant to pitch, i.e., that have similar tuning properties translated across log-frequency space \citep{kell2018task,kell2019deep} 
and possibly other sensory neurons such as somatosensory populations \citep{lieber2019high}.
But it also holds true for higher cognitive neurons, such as the ones discussed above, where, for instance, each grid cell is tuned to a different location on the torus and the receptive fields (tuning curves) across grid cells are remarkably similar in shape \citep{hafting2005microstructure,fyhn2007hippocampal,gardner2022toroidal}.
Thus, we propose to introduce \emph{feature sharing} also in this case.
More precisely, we argue that every $f_i$ within an ensemble $I_j$, should be modeled by a shared tuning curve function $g_j(z, \theta_i) = f_i(z)$ that applies a simple neuron specific transformation, parameterized by $\theta_i$, to obtain the specific $f_i$ of each neuron.
For instance, in the case of a neuron-specific spatial translation ($\mu_i$, the receptive field center as above) and scaling ($\alpha_i$), we would define $\theta_i = \{\mu_i, \alpha_i\}$ and thus $g_j(z, \theta_i) = \alpha_i g_j(T_{\mu_i}(z), \theta_0) = f_i(z)$ ($\theta_0$~yielding simply the centered prototype of the shared tuning curve).
Evidence for this~assumption is provided in Fig. \ref{Why_feature_sharing_is_a_good_idea}, where centering and rescaling all tuning curves of head direction selective neurons on $\mathbb{S}^1$ \citep[recorded by][]{peyrache2015rec} clearly exhibits a shared, Gaussian-like tuning.

In visual perception there is translational equivariance, however, only among cells of the same types \citep{wassle1978mosaic}.
Therefore, previous work proposed feature sharing in combination with functional cell type identification \citep{klindt2017neural}.
Specifically, equating neural ensembles with cell types in the retina, this posits that all neurons in a given ensemble $I_j$ be tuned to the same set of latent variables $J_j$, i.e., $J_i = J_j, \, \forall i \in I_j$.
As an example, the toroidal structures $z_j \in \mathbb{T}^2$ of different grid cell ensembles $I_j$ only become apparent after the successful separation of distinct modules with different spatial resolutions \citep{gardner2022toroidal}.
Thus, we argue that feature sharing hinges on the successful identification and separation of distinct neural ensembles in a mixed population recording.

The problem of \emph{ensemble detection} can be formalized as partitioning $n$ recorded neurons $I = \bigcup_j^k I_j$ into $k$ non-empty subsets $I_j \cap I_{j'} = \emptyset, \, \forall j \neq {j'}$.
Unfortunately, this is a combinatorial problem with $k^n$ possible states. 
Even for a `simple' problem such as clustering neurons into $k=2$ groups, this already exceeds the number of atoms in the universe ($10^{82}$) at $n \geq 273$.
With modern neuroscience routinely yielding thousands of recorded neurons per experiment \citep{jun2017fully}, brute-force exhaustive search is clearly not a viable approach.
To mitigate this issue, we propose a soft-relaxation of the clustering problem.
Specifically, we define a convex combination of weights over responses derived from tuning curves on each latent variable, and train the neurons to maximize their likelihood given those weights in a differentiable fashion (explained in more detail in the next section).

\section{Methods}\label{sec:methods}

As above, we denote neural activity $x$ and the collections of latent variables as $z$.
We want to~learn a latent variable model $p(x, z) = p(x|z) p(z)$.
We choose a variational autoencoder \citep[VAE,][]{kingma2013auto}, which maximizes a lower bound to the data likelihood called the evidence lower bound, which necessitates the introduction of an approximation to the posterior distribution, i.e., a variational (input dependent) posterior with non-zero support across the domain, $q(z|x) > 0, \, \forall z, x$.


At test-time, we focus on a precise estimate for the latents $z$. These latents are usually inferred with the encoder that computes the approximate variational posterior $q(z|x)$. Here, we aim to further refine this estimate by performing inference based on the decoder $p(x|z)$. Technically, this can be achieved by directly performing gradient descent on the latents, with the decoder starting from the encoder estimate \citep{schott2018towards, ghosh2019resisting} (see Appx. \ref{section: hybrid_inference} for further details).

For Euclidean latent spaces $z \in \mathbb{R}^n$, we can simply rely on the standard VAE framework, 
using~a Normal ($\mathcal{N}$) posterior and prior for those dimensions.
For spherical latent spaces $\mathbb{S}^n$, we use a wrapped Normal distribution ($w\mathcal{N}$), akin to previous work \citep{falorsi2019reparameterizing,jensen2020manifold}, which satisfies the topology of the latent space.
For the relevant case of a toroidal latent space, we can use $\mathbb{T}^2 = \mathbb{S}^1 \times \mathbb{S}^1$, together with the fact that the variational posterior is, usually, assumed to factorize across dimensions.
This factorization also helps keep the equations the same when working with multiple latents (corresponding to multiple ensembles), i.e., $z=\{z_1, ..., z_k\}$.

More precisely, we assume that each latent $z_j$ lives either in Euclidean space $\mathbb{R}^{n_j}$ or on a torus~$\mathbb{T}^{n_j} = \mathbb{S}^1_1 \times \hdots \times \mathbb{S}^1_{n_j}$ (with $\mathbb{S}^1=\mathbb{T}^1$).
Further, we let our prior factorize \citep{kingma2013auto}, i.e.,
\begin{equation}
    p(z)=\prod^k_{j=1} p(z_j), \quad p(z_j) = \prod^{n_j}_{l=1} p(z_j^{(l)}), \quad z_j^{(l)} \sim \left\{\begin{array}{rl}
    \mathcal{N}(0, 1), &\hspace{-0.5em}\text{if }z_j \in \mathbb{R}^{n_j}\\
    w\mathcal{N}(0, 1), &\hspace{-0.5em}\text{if }z_j \in \mathbb{T}^{n_j}\end{array}\right., \\
\end{equation}
giving uniform prior distributions.
Analogously, we let our variational posteriors factorize as
\begin{equation}
    \hspace{-2.8em}q(z|x)=\prod^k_{j=1} q(z_j|x), \quad q(z_j|x) = \prod^{n_j}_{l=1} q(z_j^{(l)}|x), \quad z_j^{(l)} \sim \left\{\begin{array}{rl}
    \mathcal{N}(\mu(x), \sigma(x)), &\hspace{-0.5em}\text{if }z_j \in \mathbb{R}^{n_j}\\
    w\mathcal{N}(\mu(x), \sigma(x)), &\hspace{-0.5em}\text{if }z_j \in \mathbb{T}^{n_j}\end{array}\right..\hspace{-1.5em} 
\end{equation}
Here, the distribution-specific parameters $\mu(x)$ and $\sigma(x)$ are themselves input dependent, and given by a functional mapping which is learned using the reparametrization trick. 
We use a temporal (1D) convolutional neural network to parameterize $q$.
To enforce circular latents, we let the encoder output vectors in $\mathbb{R}^2$, which we subsequently normalize to avoid discontinuities when estimating angles \citep{zhou2019continuity}. We remark that although a convolutional filter was used, temporal smoothness was not explicitly included in the latent dynamics of $z$ in the model used for producing the results in Sec. \ref{sec:experiments}. It is, however, a straightforward inclusion to incorporate (see Appx. \ref{section:temporal_transition_priors}). 

For the decoder, in clear contrast to standard VAEs, we pick a simple parametric function $f$ to parameterize the reconstruction term $p(x|z)$.
As argued above, this is a crucial choice in the interest of defining simple biologically meaningful link functions (i.e., tuning curves) between (potentially complex) latent variables and neural activity. Specifically, we have
\begin{equation}
    p(x_i|z) \sim \mathcal{P}(f_i(z)), \qquad  f_i(z) = \sum_j w_{ij} g_j(z_j, \theta_i),
\end{equation}
where $\mathcal{P}(f_i(z))$ is a Poisson distribution with rate $f_i(z)$, and the weights for each neuron $i$ the output of a softmax function, such that $w_{ij} > 0$ and $\sum_j w_{ij} = 1$. 
That is, $w_{ij}$ models the participation of neuron $i$ in ensemble $j$.
The function $g_j(z_j, \theta_i) \in \mathbb{R}$ characterizes the (shared) tuning curves of neurons in the latent space of $z_j$, and takes as input the corresponding latent $z_j$ and a (set of) neuron specific parameter(s) $\theta_i$, to produce the response of the neuron based on each latent space.

The function $g_j(z_j, \theta_i)$ can be specified in a multitude of ways, while still embodying the concept of feature sharing. Here we present two possible options: the first simply assumes a single heat kernel shape (i.e., Gaussian-like bump) with shared tuning width $\sigma$ for all neurons.
Since neural activity is non-negative, we model the $\log$ of the instantaneous firing rates as
\begin{equation}
    \log g_j(z_j, \mu_i) = - \frac{d_j(z_j,\mu_i)}{\sigma^2}^2,
\label{Bump_tuning}
\end{equation}
where $\mu_i$ denotes (as above) the center of the $i$-th neuron's tuning curve and $d_j(\cdot)$ a distance function in the space of $z_j$ (i.e., Euclidean distance in $\mathbb{R}^n$ and geodesic distance in $\mathbb{T}^n$).
The second, and more flexible, feature space consists of a sum of $M$ weighted ($\beta_m$) Gaussian basis functions (i.e., a spline)
\begin{equation}
    \log g_j(z_j, \mu_i) =  \sum_m \beta_{m} \exp \left[ - \frac{d_j(z_j,\mu_m -\mu_i)^2}{\sigma_m^2}\right],
\label{Splines_tuning}
\end{equation}
with center $\mu_m$ and width $\sigma_m$ of the $m$-th basis function. 
We consider three variations of~faeLVM: sharing of features by utilizing Eq.~\ref{Bump_tuning} to model $g_j$ with a shared (bump) heat kernel (faeLVM-b), sharing of features through Eq.~\ref{Splines_tuning} using a Gaussian basis (faeLVM-s), as well as a model allowing each neuron to learn its own set of the basis weights in Eq.~\ref{Splines_tuning}, i.e. no shared features (faeLVM-n).

\section{Experiments}\label{sec:experiments}

\begin{figure}
\floatbox[{\capbeside\thisfloatsetup{capbesideposition={right,center},capbesidewidth=0.62\textwidth}}]{figure}[\FBwidth]
{\caption{\textbf{Schematic outline of model evaluation}. Evaluation pipeline as suggested in  \citet{PeiYe2021NeuralLatents}. During training, the models are all trained on spikes from `train' neurons \textbf{(}$\mathbf{1}$\textbf{)}, learning their tuning curves, and the latent variable representation which is used to infer tuning curves on `test' neurons at train time \textbf{(}$\mathbf{2}$\textbf{)}. For testing, we fix all tuning curve shapes and infer the latents from test data \textbf{(}$\mathbf{3}$\textbf{)}, before performing spike prediction on held-out data from held-out neurons \textbf{(}$\mathbf{4}$\textbf{)}. Thus, our encoder only receives train neurons ($\mathbf{1}$, $\mathbf{3}$) to infer latents.}\label{Train_Test}}
{\includegraphics[width=0.35\textwidth]{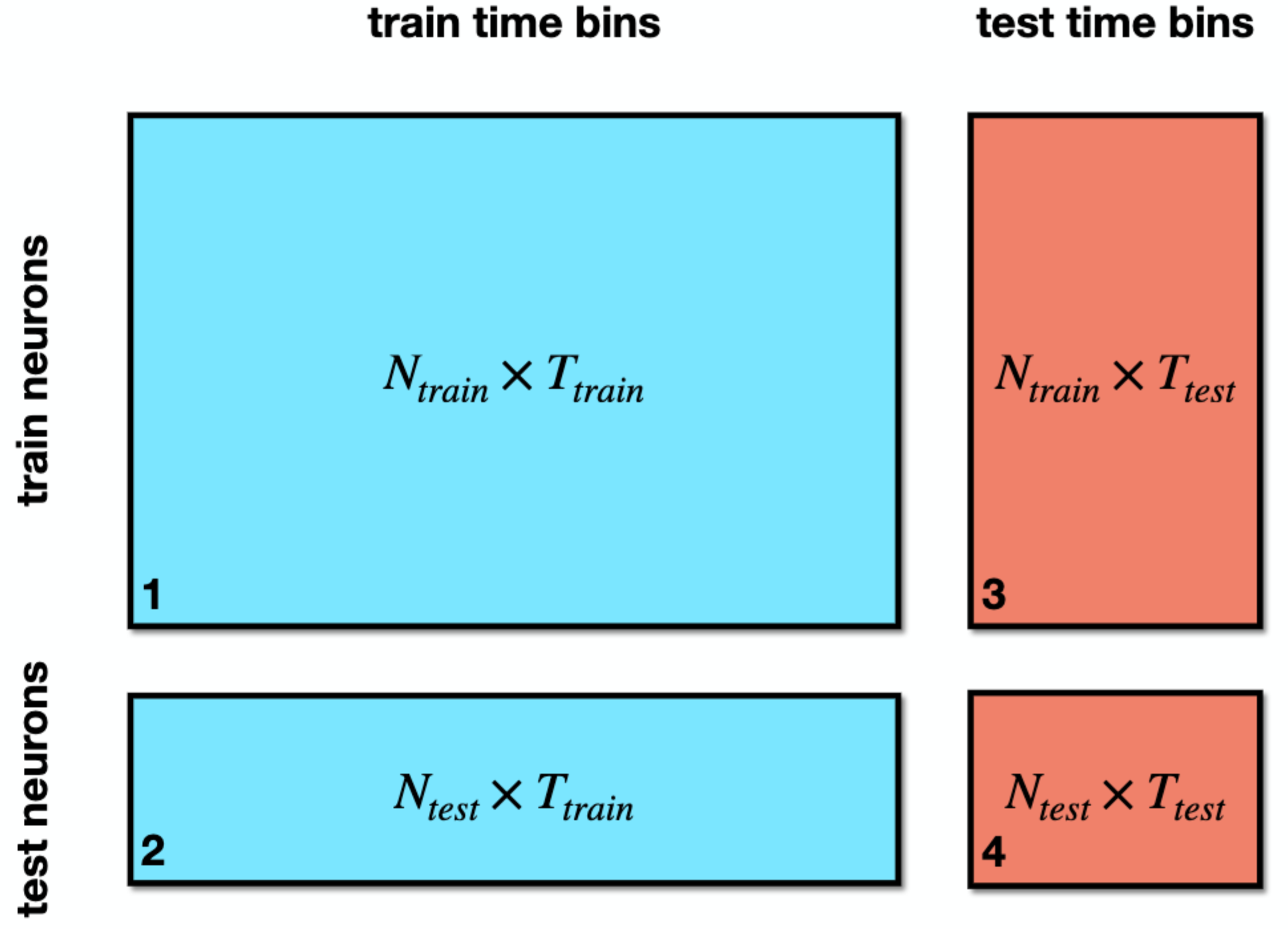}}
\end{figure}

\subsection{Simulations: Feature Sharing}
\label{sec:feature_sharing_sims}

Motivated by the evaluation pipeline in \citet{PeiYe2021NeuralLatents}, we split our data into four parts (Fig.~\ref{Train_Test}), exploring the benefits of feature sharing in the following way: We generate $z \in \mathbb{S}^1$, a circular 1D latent variable and simulate spikes according to a Poisson distribution with heat kernel tuning curves (see also Appx. \ref{section:calcium_visual_example} for a retinal ganglion simulation with calcium dynamics). 
We fix $T_{\text{test}}$ and $N_{\text{test}}$ ($1000$ and $30$), and investigate model performance based on fixed $T_{\text{train}}$ and varying $N_{\text{train}}$, and vice versa. For each condition, we repeat the experiment over $20$ seeds (while data is different for each repeat, all models are trained on the same seed, ensuring valid comparisons), reporting mean negative log-likelihood (NLLH) and geodesic error (GE) on test data (lower numbers are better, in both cases). We also compare our models with mGPLVM \citep{jensen2020manifold}, an extension of GPLVM \citep{wu2018learning} designed for inference on non-Euclidean spaces, as well as a conventional VAE using a standard multi-layer perceptron decoder (MLP; with essentially a reverse architecture of the encoder). As mGPLVM assumes a Gaussian noise model, the Poisson NLLH is unsuitable as a measure, hence it is only included when evaluating latent recoveries.

\begin{figure}[t]
  \centering
  \includegraphics[width=1\textwidth]{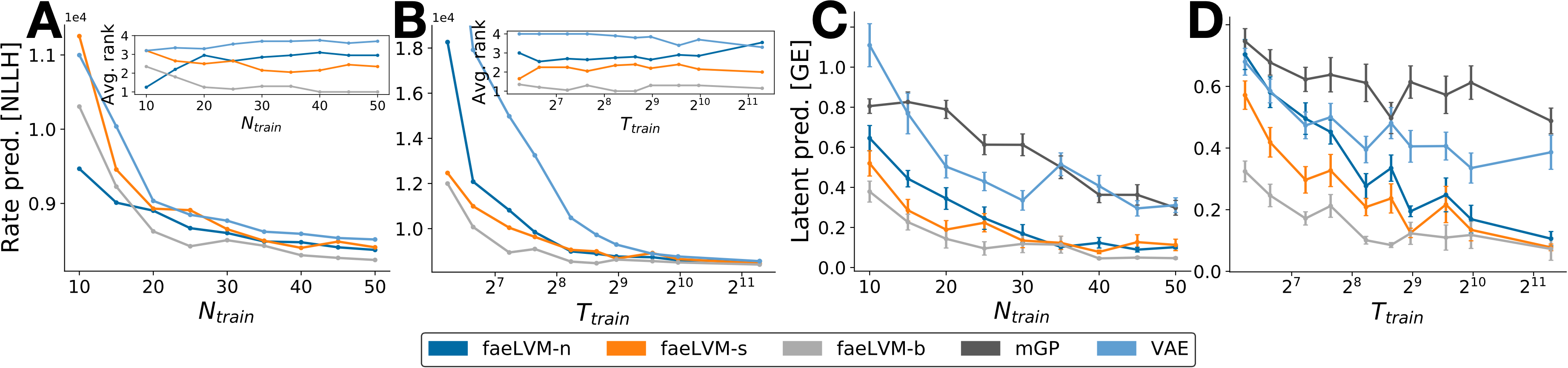}
  \caption{\textbf{Latent and Rate Prediction}. The models faeLVM-{n,s,b} are described in the main text, mGP is the model from \citet{jensen2020manifold}, VAE a traditional approach using an MLP decoder. \textbf{A},~Mean NLLH of predicted test neuron rate, as a function of amount of training neurons. As the performance is averaged over $20$ seeds of varying data, error bars would also reflect the variability in NLLH across data, resulting in inappropriate visualization. Hence, we report the mean rank of $20$ seeds for each model at each condition, in an effort to more appropriately capture the variability of the models. \textbf{B}, Same as A, now as a function of training time points. \textbf{C}, Mean GE between true latent and inferred latent, on test data, with corresponding SEM error bars, as a function of number of training neurons. \textbf{D}, Same as C, now as a function of training time points.}
  \label{Scaling_in_T_and_N}
\end{figure}

Fig. \ref{Scaling_in_T_and_N}A-B clearly show the regime where feature sharing excels: datasets with a satisfactory number of neurons, but with short recording times. This coincides well with the current state-of-affairs for neural recording techniques, where experiments are usually limited by the possible experimental length, not by the amount of recordable neurons \citep{stevenson2011advances}. Moreover, even with further technical innovations, we may eventually be able to record from all neurons, but we will likely never be able to record during all possible natural inputs or behaviors.

We also remark that while faeLVM-n seems to performs better on rate prediction in the low neuron regime (Fig. \ref{Scaling_in_T_and_N}A), the performance of LVMs ultimately hinges on how well the models can recover the correct latent variable. In the setting of latent recovery (Fig. \ref{Scaling_in_T_and_N}C-D), we see that feature sharing is evidently beneficial both in low neuron- and low recording-time regimes. faeLVM-b outperforms the other models, perhaps somewhat expected given the shape of the synthetic tuning curves, while all faeLVMs outperform both mGPLVM and VAE. We also observe that while mGPLVM, which uses tuning curve decoders, is more interpretable, it is worse than neural network decoders. However, our approach with feature sharing closes that gap and even improves upon the existing VAE method, thus highlighting the importance of sharing features; more accurate inference, while also being interpretable.
Note that although the faeLVMs leverage the benefit of a convolutional layer (which mGPLVM does not, being a non-convolutional model), we emphasize that the faeLVMs still outperforms mGPLVM, even after reducing the convolutional filter size down to $1$, which forces the model to infer the latents independently at each time point (see Appx. \ref{section: ablation_conv_filter}).

\subsection{Simulations: Ensemble Detection}
\label{sec:ensemble_detection}

\begin{figure}[h]
  \centering
  \includegraphics[width=1\textwidth]{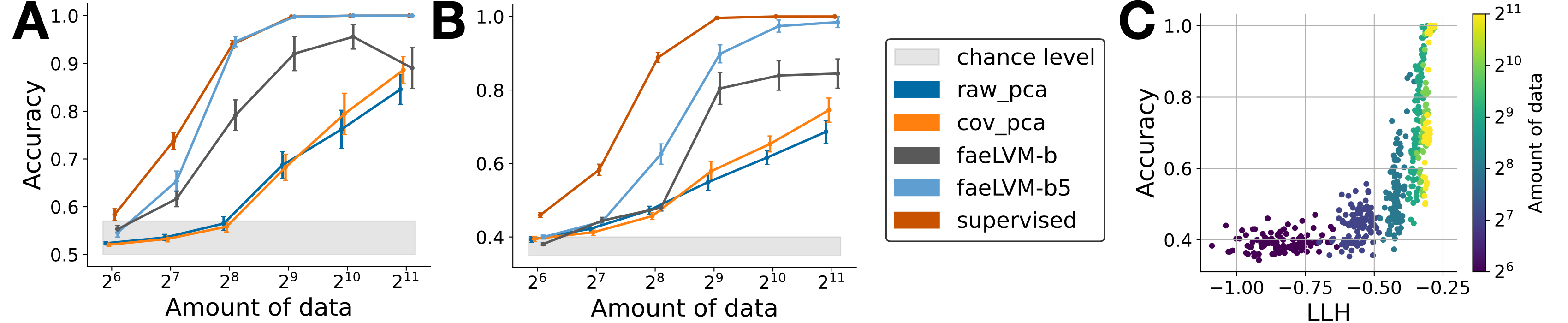}
  \caption{\textbf{Ensemble Detection}. Ensemble detection accuracy, with corresponding SEM error bars, on two (\textbf{A}) and three (\textbf{B}) equally sized ensembles of neurons tuned to separate $\mathbb{S}^1$ latent variables. The models are: k-means on PCA reduced data (`raw-pca'), k-means on PCA reduced covariance matrix of the data (`cov-pca'), supervised clustering based on the mutual information scores between the cell responses and the latent variables, faeLVM with shared heat kernel tuning curves (`faeLVM-b'), and the same model optimized over 5 seeds, selecting the one that produced the highest LLH (`faeLVM-b5'). Chance levels are computed from $100,000$ random labels with optimal permutations. \textbf{C},~Relation between likelihood and accuracy for different data amounts.}
  \label{ensemble_detection}
  \vspace{10pt}
\end{figure}

Next, we shift our focus to the challenge of ensemble detection, comparing the performance of faeLVM-b against other common methods.
Specifically, we contrast against clustering methods (see Appx. \ref{section:ensemble_detection_comparisons_explained} for more details) that either i) perform k-means clustering on dimensionally reduced neural activity over time \citep[loosely inspired by][]{lopes2011neuronal,baden2016functional,hamm2021cortical}, or ii) perform clustering on the neural covariance matrix \citep[loosely inspired by][]{carrillo2015endogenous}.
As an additional upper bound to the achievable performance, we include a \emph{supervised} clustering method that receives both the mutual information score 
between each neuron's activity, and the true latent variables, as input.
We run our model over five different random seeds and pick the run that yields the highest LLH (faeLVM-b5 in Fig. \ref{ensemble_detection}), a fair approach as the likelihood is an unsupervised metric (k-means is run over $100$ seeds, picking the one with the lowest inertia).

The setting we chose for these experiments was the separation of ensembles of neurons that live on~a ring, i.e. $\mathbb{S}^1$.
The comparison methods failed to separate the ensembles when scaling up to neural activity on a torus, in contrast to faeLVM, which further highlights the novelty of an unsupervised method that can accomplish this task (see Sec. \ref{sec:grid cells data} for tori, and also Appx. \ref{section:additional_ensembles} for an additional ensemble separation ablation).
We tested the accuracy on separating two~and three distinct (no shared latents across ensembles) neural ensembles of equal size on rings, in addition to investigating accuracy for all models as a function of available data~points (recording~length).

Fig. \ref{ensemble_detection}A-B shows that our model outperforms both comparison methods and quickly approaches the performance of the supervised upper bound, for both two and three rings.
This is even more pronounced for the model selected based on LLH --- a viable option when the analysis result warrants additional computing time.
Fig. \ref{ensemble_detection}C shows that the relationship between ensemble detection accuracy and LLH is monotonically increasing.
Thus, the unsupervised likelihood score is indeed a good proxy to arbitrate between different optimization runs.

\subsection{Real Datasets: Head Direction Data}\label{sec:head_cells}

\begin{figure}[h]
  \centering
  \includegraphics[width=1\textwidth]{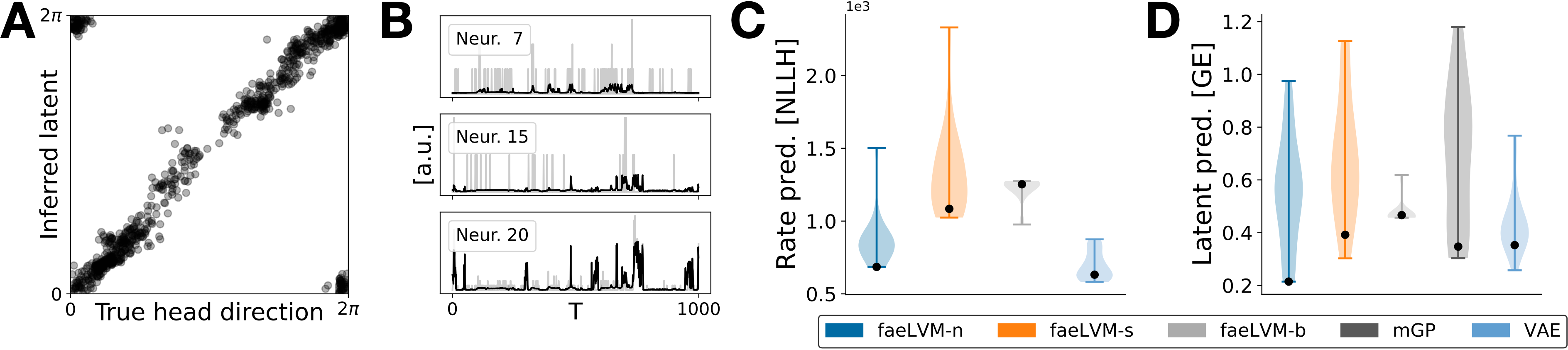}
  \caption{\textbf{Application to head direction data}. \textbf{A}, Variational mean of faeLVM-n plotted against recorded mouse head direction (for additional visualizations, see Appx. \ref{section:additional_visualizations_real_data}). We trained $20$ models, selecting the one with highest training likelihood on training neurons, before performing inference at test-time. \textbf{B}, True test neuron spikes (grey), and predicted test neuron rates (black) from the model used in A, per bin, for three randomly selected neurons. \textbf{C}, NLLH of test neuron rate, over $20$ seeds, as a violin plot.
  Performance of the best model, based on train LLH, is indicated by the black dot. \textbf{D}, Same as C, but comparing GE between recorded head direction and inferred latent, on test data.\vspace{5pt}}
  \label{Application_to_peyrache}
\end{figure}

Transitioning to real datasets, we first apply our models to data from \citet{peyrache2015rec}
; more specifically the awake trials from the dataset labeled Mouse $28$, session $140313$, recorded from the anterodorsal thalamic nucleus (ADN). Recordings yield approx. $35$ min. of data, which we bin in $100$ ms bins, before separating it with a train-test split of $95\% - 5\%$. $3$ neurons are allocated as test neurons (out of $26$ total) and we run $20$ seeds of each model, selecting the best performing model for visualization (faeLVM-n). The inferred latent (Fig. \ref{Application_to_peyrache}A) is observed to match the true head direction to a high degree, and the model is able to predict rates on test neurons (Fig. \ref{Application_to_peyrache}B), even though the average and peak activity over test bins are quite low ($\sim1$ and $5$ Hz respectively) for the first two.

As we can see from Fig. \ref{Application_to_peyrache}C, the unconstrained model (faeLVM-n) outperforms the other two. This is not an unsurprising result, given the model's higher complexity and the sufficiently large amount of training data. As for the latent prediction, mGPLVM and the traditional VAE (Fig. \ref{Application_to_peyrache}D) are close in comparison to the performance of faeLVM-s, although less accurate than faeLVM-n. We also note that the VAE does perform better on the rate prediction, which, conceivably, might be explained by the fact that it is not constrained by a particular tuning curve model and thus might learn interactions that incorporate non-head direction variables in an effort to improve its rate prediction. Training multiple models and making a selection based on train LLH generally correlates well with solid results on the test set, as also shown in Fig \ref{ensemble_detection}C.
Note that faeLVM-b achieves much more consistent optimization results over different random initializations, suggesting a typical bias-variance tradeoff.

Overall, we recognize that while there are differences in performances, all models recover the correct circular manifold. The faeLVMs however, are noticeably faster ($\sim 5$ min. run time for one seed, on a CPU), compared to mGPLVM ($\sim 50$ min. run time for one seed on a GPU). While not critical in the case of this particular dataset, it is paramount when inferring latents of higher dimensionality.
We also note that while the circular manifold was in this case known, it is possible to do model selection over different latent spaces (toroidal, circular, planar, etc.) as was demonstrated in \citet{jensen2020manifold}. 
Thus, it is not strictly necessary to know the latent topology when applying our method, rather it allows for the restriction of possible models to a limited hypothesis class.

\subsection{Real Datasets: Grid Cell Data}
\label{sec:grid cells data}

\begin{figure}[t]
    \centering
    \includegraphics[width=1\textwidth]{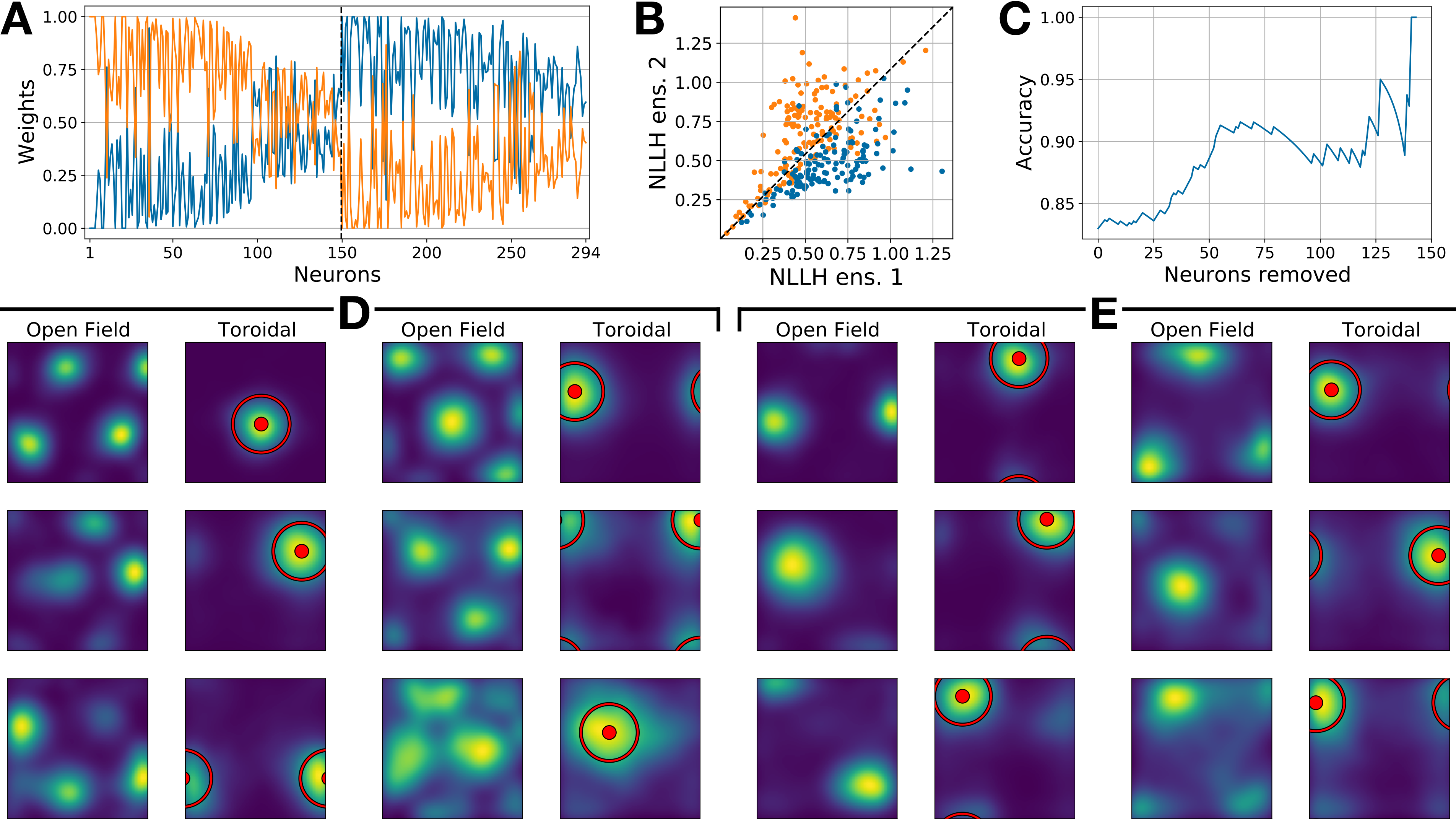}
    \caption{\textbf{Application to grid cell data}. \textbf{A}, Learned ensemble weights. Neurons are ordered by decreasing spatial information, dashed line separates first and second reported module \citep{gardner2022toroidal}. \textbf{B}, Greedy test for each neuron, checking which latent provides higher likelihood (i.e., one-hot ensemble weights), colored by reported module. \textbf{C}, Classification accuracy when successively removing neurons with low spatial selectivity. The same number of neurons were removed from each ensemble, starting with least spatially informative ones. Accuracy is measured against supervised ensemble detection from \citet{gardner2022toroidal}. \textbf{D}, Rate map of every $25^{\text{th}}$ neuron, first grid cell module (smoothed for visualization purposes). Maps shown as function of rat's position in arena, as well as inferred coordinates on latent torus. Red circles indicate inferred receptive field mean and width (SD) for the heat kernel (i.e., Gaussian-like bump). \textbf{E}, Same as D for second module. For a full set of rate maps, as well as  latent dynamics, see Appx. \ref{section: all_grid_cell_maps} \& \ref{section:additional_visualizations_real_data}. \vspace{-5pt}}
    \label{Application_to_gridcells}
\end{figure}

Finally, we apply the complete model pipeline to data recorded from the medial entorhinal cortex (MEC) of a freely moving rat \citep{gardner2022toroidal}. Neurons in the MEC are selective to a number of features like space \citep{fyhn2004spatial}, head direction \citep{sargolini2006conjunctive}, speed \citep{kropff2015speed}, as well as conjunctive representations of the three \citep{sargolini2006conjunctive, hardcastle2017multiplexed}, though most demonstrate a strong preference for single features \citep{kropff2015speed, tukker2022microcircuits}. We consider data from the rat R, day $1$-session, using recordings from neurons labeled module $2$ and $3$, considering $149$ and $145$ non-conjunctive neurons respectively. Spike data was binned in $100$ ms bins, after removing periods in time when the rat was considered stationary (speed $<2.5\text{ cm}\text{ s}^{-1}$). Recordings from both the open field and maze experiments yields approx. $170$~min. of data. 
We train $20$ models from different seeds, selecting the one with highest in-sample~likelihood.

Compared with the modules discovered in the dataset \citep{gardner2022toroidal}, our model's results coincides with an accuracy of $0.83$. The inferred weights (Fig. \ref{Application_to_gridcells}A) show a clear separation between the two modules, becoming less distinct as the neurons' spatial information decreases. This is also reflected in Fig \ref{Application_to_gridcells}B, where neurons further away from the diagonal tend to be more accurately classified. However, we note that the original separation of neurons into ensembles is not an intrinsic truth; a number of the less informative neurons exhibit structurally ambiguous rate maps (see Appx. \ref{section: all_grid_cell_maps}), a possible indication that they may belong to a different ensemble altogether. Excluding these neurons while evaluating the model (Fig. \ref{Application_to_gridcells}C), we observe a significant increase in performance, reaching $0.90$ accuracy after removing approx. $50$ of the least informative neurons from each ensemble. 

Fig. \ref{Application_to_gridcells}D-E show a selection of rate maps for neurons from both ensembles. We see that the model infers toroidal decodings that keep the structural integrity of the receptive fields, akin to the findings in \citet{gardner2022toroidal}. The inferred centers and widths also correspond well with the rate maps. We note that, while promising, the dataset included here contains just $294$ of the $2460$ neurons originally recorded and that the full dataset (currently not available) would require additional steps~to determine the number of ensembles, the corresponding shapes and allow for some degree of conjunctivity. 

\section{Conclusion}\label{sec:clf}
In this paper we provide a proof-of-concept that the paradigm of share and divide is a successful application of the old idea of \emph{carving nature at the joints},\footnote{Plato (428-348 BC), \emph{Phaedrus} 265e; also, Zhuangzi (369-286 BC), \emph{Nourishing the Lord of Life}.} in the sense that similar observations are grouped while differences are highlighted.
This principle, instantiated in a group equivariant computation core and a cell-type specific readout, has been successfully deployed in visual neuroscience; we show that it also provides an excellent inductive bias (or prior) in the case of higher cognitive variables such as (but not necessarily limited to) head direction or position in space.
It is advantageous from a statistical perspective to leverage the similarity in tuning functions across neurons, as well as from a biological point of view, as finding protoypical tuning curves can help us unravel the common tuning properties across individual neurons.

The idea of \emph{separating ensembles} is at the core of the scientific endeavor where the task is to group our observations of nature into distinct categories.
In the case of grid cell modules, it is absolutely crucial, as the topological properties that define this class of neurons are completely hidden when the separate ensembles of different spatial resolution are not properly distinguished.
One might wonder whether other brain areas such as primary visual cortex, for which a power law in the population activation space has been stipulated \citep{stringer2019high}, may not resolve into simpler underlying topological ensembles of neurons, such as the special orthogonal group of three-dimensional rotations $SO(3)$ (vision), 2-spheres $\mathbb{S}^2$ \citep{singh2008topological}, or the Klein bottle shape of oriented and phase shifted wavelets \citep{carlsson2008local} (relevant to multiple sensory modalities).

Furthermore, the assumption of \emph{feature sharing} relies on an equivariant neural code, such as the approximate translation equivariance in the visual system, which certainly is an abstraction from the actual inter-neural variability that exists in the real world.
In some cases, it might be that neurons are better described along some continuous space of variations, rather than by a fixed set of discrete prototypical tuning curves \citep{ustyuzhaninov2022digital}.
On the other hand, if we can identify the parameters (e.g., tuning width) that describe the continuous variations across neurons and separate them from the stable features of tuning curves (e.g., Gaussian-like heat kernel shape), we can use that to build a more flexible parametric (but still shared) feature basis.
Examples of such approaches include rotation \citep{ustyuzhaninov2019rotation}, kinetic \citep{zhao2020temporal} or more general nonlinear modifications \citep{shah2022individual} to shared feature spaces.
Further challenges to the feature sharing idea could arise from recent findings of representational drift \citep{rule2019causes} and location- \citep{qiu2020mouse} or context dependent tuning \citep{kanter2017novel}, although future directions might consider addressing these in the model.
Other limitations are the fact that the grid cell torus is hexagonal, while we use an orthogonal basis; further analyses might reveal if the encoder compensates for that, e.g., by adjusting the distribution of speeds on the torus per direction.

Finally, although the choice of variational posteriors might at first seem restrictive, we would argue that the corresponding topologies (circles, tori, $\R^n$) are natural choices for latent spaces and likely to keep appearing in the brain, as is suggested in recent work \citep{kriegeskorte2021neural}. 
Other authors conjectured that, e.g., prefrontal cortex deploys grid cell-like codes to explore physical \citep{doeller2010evidence, jacobs2013direct} and conceptual spaces \citep{constantinescu2016organizing, bao2019grid}, as well as rules in a reinforcement learning task \citep{baram2021entorhinal}.
The challenge of constructing or observing the relevant space has proven a hurdle for previous supervised methods, being unable to discover cognitive maps without \emph{a priori} knowledge of the represented covariates.
Our model is particularly suited in this setting, offering a targeted method to identify neural ensembles and summarizing as well as leveraging their shared tuning properties.


\newpage
\subsubsection*{Ethics Statement}

More generally, reducing high dimensional nonlinear dynamical systems to a set of latent variables can have a broad array of possible applications, for instance, in climate sciences, crop forecasting or flood prediction.
Restraining the decoders to interpretable mechanisms and groups of observations could, also in those cases, help us further our understanding of the underlying processes or relations between low dimensional summaries and data observations.
As for any complex and versatile modern machine learning method, there also exists a danger of translating the insights from this work to detrimental purposes and intents.
Applications to sensitive data with protected attributes, such as gender or ethnicity, should pay special attention to the en- and decoding of those attributes in the learned latent spaces.
Therefore, while we do not see any obvious misuse, nor want to explicitly name any possible malicious purposes, we still strongly discourage any nefarious applications of the ideas developed in this work.
Lastly, we tried to minimize the environmental impact of our research by performing the hyper-parameter search on the GPU cluster of a carbon-neutral organization.

\subsubsection*{Reproducibility Statement}

Code and implementation are included as supplementary material, and is also made available at: \href{https://github.com/david-klindt/NeuralLVM}{https://github.com/david-klindt/NeuralLVM}. Datasets are public, and details regarding model parameters, data generation, etc. are specified under the relevant sections (e.g., Sec. \ref{sec:experiments}), as well as more thoroughly in the Appendix (e.g., Appx. \ref{section:appendix_model_details}, \ref{section:data_generation_for_sims}, \ref{section:ensemble_detection_comparisons_explained}, \ref{section: hybrid_inference} \& \ref{section: hyperparams}).


\subsubsection*{Acknowledgments}

We thank Erik Hermansen for valuable discussions and feedback related to the grid cell data, as well as Ta-Chu Kao for fruitful discussions. This work was supported by a Norwegian Research Council Large-scale Interdisciplinary Researcher Project Grant iMOD (NFR grant no. 325114).

\bibliography{iclr2023_conference}

\begin{thebibliography}{101}
\providecommand{\natexlab}[1]{#1}
\providecommand{\url}[1]{\texttt{#1}}
\expandafter\ifx\csname urlstyle\endcsname\relax
  \providecommand{\doi}[1]{doi: #1}\else
  \providecommand{\doi}{doi: \begingroup \urlstyle{rm}\Url}\fi

\bibitem[Albright(1984)]{albright1984direction}
Thomas~D Albright.
\newblock Direction and orientation selectivity of neurons in visual area mt of
  the macaque.
\newblock \emph{Journal of neurophysiology}, 52\penalty0 (6):\penalty0
  1106--1130, 1984.

\bibitem[Baden et~al.(2016)Baden, Berens, Franke, Rom{\'a}n~Ros{\'o}n, Bethge,
  and Euler]{baden2016functional}
Tom Baden, Philipp Berens, Katrin Franke, Miroslav Rom{\'a}n~Ros{\'o}n,
  Matthias Bethge, and Thomas Euler.
\newblock The functional diversity of retinal ganglion cells in the mouse.
\newblock \emph{Nature}, 529\penalty0 (7586):\penalty0 345--350, 2016.

\bibitem[Bao et~al.(2019)Bao, Gjorgieva, Shanahan, Howard, Kahnt, and
  Gottfried]{bao2019grid}
Xiaojun Bao, Eva Gjorgieva, Laura~K Shanahan, James~D Howard, Thorsten Kahnt,
  and Jay~A Gottfried.
\newblock Grid-like neural representations support olfactory navigation of a
  two-dimensional odor space.
\newblock \emph{Neuron}, 102\penalty0 (5):\penalty0 1066--1075, 2019.

\bibitem[Baram et~al.(2021)Baram, Muller, Nili, Garvert, and
  Behrens]{baram2021entorhinal}
Alon~Boaz Baram, Timothy~Howard Muller, Hamed Nili, Mona~Maria Garvert, and
  Timothy Edward~John Behrens.
\newblock Entorhinal and ventromedial prefrontal cortices abstract and
  generalize the structure of reinforcement learning problems.
\newblock \emph{Neuron}, 109\penalty0 (4):\penalty0 713--723, 2021.

\bibitem[Bashiri et~al.(2021)Bashiri, Walker, Lurz, Jagadish, Muhammad, Ding,
  Ding, Tolias, and Sinz]{bashiri2021flow}
Mohammad Bashiri, Edgar Walker, Konstantin-Klemens Lurz, Akshay Jagadish,
  Taliah Muhammad, Zhiwei Ding, Zhuokun Ding, Andreas Tolias, and Fabian Sinz.
\newblock A flow-based latent state generative model of neural population
  responses to natural images.
\newblock \emph{Advances in Neural Information Processing Systems}, 34, 2021.

\bibitem[Batty et~al.(2017)Batty, Merel, Brackbill, Heitman, Sher, Litke,
  Chichilnisky, and Paninski]{batty2017multilayer}
Eleanor Batty, Josh Merel, Nora Brackbill, Alexander Heitman, Alexander Sher,
  Alan Litke, EJ~Chichilnisky, and Liam Paninski.
\newblock Multilayer recurrent network models of primate retinal ganglion cell
  responses.
\newblock In \emph{5th International Conference on Learning Representations},
  2017.

\bibitem[Bergstra \& Bengio(2012)Bergstra and Bengio]{bergstra2012random}
James Bergstra and Yoshua Bengio.
\newblock Random search for hyper-parameter optimization.
\newblock \emph{Journal of machine learning research}, 13\penalty0 (2), 2012.

\bibitem[Burg et~al.(2021)Burg, Cadena, Denfield, Walker, Tolias, Bethge, and
  Ecker]{burg2021learning}
Max~F Burg, Santiago~A Cadena, George~H Denfield, Edgar~Y Walker, Andreas~S
  Tolias, Matthias Bethge, and Alexander~S Ecker.
\newblock Learning divisive normalization in primary visual cortex.
\newblock \emph{PLOS Computational Biology}, 17\penalty0 (6):\penalty0
  e1009028, 2021.

\bibitem[Cadena et~al.(2019)Cadena, Sinz, Muhammad, Froudarakis, Cobos, Walker,
  Reimer, Bethge, Tolias, and Ecker]{cadena2019well}
Santiago~A Cadena, Fabian~H Sinz, Taliah Muhammad, Emmanouil Froudarakis, Erick
  Cobos, Edgar~Y Walker, Jake Reimer, Matthias Bethge, Andreas Tolias, and
  Alexander~S Ecker.
\newblock How well do deep neural networks trained on object recognition
  characterize the mouse visual system?
\newblock 2019.

\bibitem[Carlsson et~al.(2008)Carlsson, Ishkhanov, De~Silva, and
  Zomorodian]{carlsson2008local}
Gunnar Carlsson, Tigran Ishkhanov, Vin De~Silva, and Afra Zomorodian.
\newblock On the local behavior of spaces of natural images.
\newblock \emph{International journal of computer vision}, 76\penalty0
  (1):\penalty0 1--12, 2008.

\bibitem[Carrillo-Reid et~al.(2015)Carrillo-Reid, Miller, Hamm, Jackson, and
  Yuste]{carrillo2015endogenous}
Luis Carrillo-Reid, Jae-eun~Kang Miller, Jordan~P Hamm, Jesse Jackson, and
  Rafael Yuste.
\newblock Endogenous sequential cortical activity evoked by visual stimuli.
\newblock \emph{Journal of neuroscience}, 35\penalty0 (23):\penalty0
  8813--8828, 2015.

\bibitem[Chaudhuri et~al.(2019)Chaudhuri, Ger{\c{c}}ek, Pandey, Peyrache, and
  Fiete]{chaudhuri2019intrinsic}
Rishidev Chaudhuri, Berk Ger{\c{c}}ek, Biraj Pandey, Adrien Peyrache, and Ila
  Fiete.
\newblock The intrinsic attractor manifold and population dynamics of a
  canonical cognitive circuit across waking and sleep.
\newblock \emph{Nature neuroscience}, 22\penalty0 (9):\penalty0 1512--1520,
  2019.

\bibitem[Churchland et~al.(2012)Churchland, Cunningham, Kaufman, Foster,
  Nuyujukian, Ryu, and Shenoy]{churchland2012neural}
Mark~M Churchland, John~P Cunningham, Matthew~T Kaufman, Justin~D Foster, Paul
  Nuyujukian, Stephen~I Ryu, and Krishna~V Shenoy.
\newblock Neural population dynamics during reaching.
\newblock \emph{Nature}, 487\penalty0 (7405):\penalty0 51--56, 2012.

\bibitem[Constantinescu et~al.(2016)Constantinescu, O’Reilly, and
  Behrens]{constantinescu2016organizing}
Alexandra~O Constantinescu, Jill~X O’Reilly, and Timothy~EJ Behrens.
\newblock Organizing conceptual knowledge in humans with a gridlike code.
\newblock \emph{Science}, 352\penalty0 (6292):\penalty0 1464--1468, 2016.

\bibitem[Cotton et~al.(2020)Cotton, Sinz, and Tolias]{cotton2020factorized}
Ronald~James Cotton, Fabian Sinz, and Andreas Tolias.
\newblock Factorized neural processes for neural processes: K-shot prediction
  of neural responses.
\newblock \emph{Advances in Neural Information Processing Systems},
  33:\penalty0 11368--11379, 2020.

\bibitem[Curto(2017)]{curto2017can}
Carina Curto.
\newblock What can topology tell us about the neural code?
\newblock \emph{Bulletin of the American Mathematical Society}, 54\penalty0
  (1):\penalty0 63--78, 2017.

\bibitem[Davidson et~al.(2018)Davidson, Falorsi, De~Cao, Kipf, and
  Tomczak]{davidson2018hyperspherical}
Tim~R Davidson, Luca Falorsi, Nicola De~Cao, Thomas Kipf, and Jakub~M Tomczak.
\newblock Hyperspherical variational auto-encoders.
\newblock \emph{arXiv preprint arXiv:1804.00891}, 2018.

\bibitem[Doeller et~al.(2010)Doeller, Barry, and Burgess]{doeller2010evidence}
Christian~F Doeller, Caswell Barry, and Neil Burgess.
\newblock Evidence for grid cells in a human memory network.
\newblock \emph{Nature}, 463\penalty0 (7281):\penalty0 657--661, 2010.

\bibitem[Ecker et~al.(2018)Ecker, Sinz, Froudarakis, Fahey, Cadena, Walker,
  Cobos, Reimer, Tolias, and Bethge]{ecker2018rotation}
Alexander~S Ecker, Fabian~H Sinz, Emmanouil Froudarakis, Paul~G Fahey,
  Santiago~A Cadena, Edgar~Y Walker, Erick Cobos, Jacob Reimer, Andreas~S
  Tolias, and Matthias Bethge.
\newblock A rotation-equivariant convolutional neural network model of primary
  visual cortex.
\newblock \emph{arXiv preprint arXiv:1809.10504}, 2018.

\bibitem[Elsayed \& Cunningham(2017)Elsayed and
  Cunningham]{elsayed2017structure}
Gamaleldin~F Elsayed and John~P Cunningham.
\newblock Structure in neural population recordings: an expected byproduct of
  simpler phenomena?
\newblock \emph{Nature neuroscience}, 20\penalty0 (9):\penalty0 1310--1318,
  2017.

\bibitem[Falorsi et~al.(2019)Falorsi, de~Haan, Davidson, and
  Forr{\'e}]{falorsi2019reparameterizing}
Luca Falorsi, Pim de~Haan, Tim~R Davidson, and Patrick Forr{\'e}.
\newblock Reparameterizing distributions on lie groups.
\newblock In \emph{The 22nd International Conference on Artificial Intelligence
  and Statistics}, pp.\  3244--3253. PMLR, 2019.

\bibitem[Franke et~al.(2021)Franke, Willeke, Ponder, Galdamez, Muhammad, Patel,
  Froudarakis, Reimer, Sinz, and Tolias]{franke2021behavioral}
Katrin Franke, Konstantin~F Willeke, Kayla Ponder, Mario Galdamez, Taliah
  Muhammad, Saumil Patel, Emmanouil Froudarakis, Jacob Reimer, Fabian Sinz, and
  Andreas Tolias.
\newblock Behavioral state tunes mouse vision to ethological features through
  pupil dilation.
\newblock \emph{bioRxiv}, 2021.

\bibitem[Fuhs \& Touretzky(2006)Fuhs and Touretzky]{fuhs2006spin}
Mark~C Fuhs and David~S Touretzky.
\newblock A spin glass model of path integration in rat medial entorhinal
  cortex.
\newblock \emph{Journal of Neuroscience}, 26\penalty0 (16):\penalty0
  4266--4276, 2006.

\bibitem[Fyhn et~al.(2004)Fyhn, Molden, Witter, Moser, and
  Moser]{fyhn2004spatial}
Marianne Fyhn, Sturla Molden, Menno~P Witter, Edvard~I Moser, and May-Britt
  Moser.
\newblock Spatial representation in the entorhinal cortex.
\newblock \emph{Science}, 305\penalty0 (5688):\penalty0 1258--1264, 2004.

\bibitem[Fyhn et~al.(2007)Fyhn, Hafting, Treves, Moser, and
  Moser]{fyhn2007hippocampal}
Marianne Fyhn, Torkel Hafting, Alessandro Treves, May-Britt Moser, and Edvard~I
  Moser.
\newblock Hippocampal remapping and grid realignment in entorhinal cortex.
\newblock \emph{Nature}, 446\penalty0 (7132):\penalty0 190--194, 2007.

\bibitem[Gallego et~al.(2017)Gallego, Perich, Miller, and
  Solla]{gallego2017neural}
Juan~A Gallego, Matthew~G Perich, Lee~E Miller, and Sara~A Solla.
\newblock Neural manifolds for the control of movement.
\newblock \emph{Neuron}, 94\penalty0 (5):\penalty0 978--984, 2017.

\bibitem[Gao et~al.(2017)Gao, Trautmann, Yu, Santhanam, Ryu, Shenoy, and
  Ganguli]{gao2017theory}
Peiran Gao, Eric Trautmann, Byron Yu, Gopal Santhanam, Stephen Ryu, Krishna
  Shenoy, and Surya Ganguli.
\newblock A theory of multineuronal dimensionality, dynamics and measurement.
\newblock \emph{BioRxiv}, pp.\  214262, 2017.

\bibitem[Gardner et~al.(2022)Gardner, Hermansen, Pachitariu, Burak, Baas, Dunn,
  Moser, and Moser]{gardner2022toroidal}
Richard~J Gardner, Erik Hermansen, Marius Pachitariu, Yoram Burak, Nils~A Baas,
  Benjamin~A Dunn, May-Britt Moser, and Edvard~I Moser.
\newblock Toroidal topology of population activity in grid cells.
\newblock \emph{Nature}, pp.\  1--6, 2022.

\bibitem[Ghosh et~al.(2019)Ghosh, Losalka, and Black]{ghosh2019resisting}
Partha Ghosh, Arpan Losalka, and Michael~J Black.
\newblock Resisting adversarial attacks using gaussian mixture variational
  autoencoders.
\newblock In \emph{Proceedings of the AAAI Conference on Artificial
  Intelligence}, volume~33, pp.\  541--548, 2019.

\bibitem[Goldin et~al.(2021)Goldin, Lefebvre, Virgili, Ecker, Mora, Ferrari,
  and Marre]{goldin2021context}
Mat{\'\i}as~A Goldin, Baptiste Lefebvre, Samuele Virgili, Alexander Ecker,
  Thierry Mora, Ulisse Ferrari, and Olivier Marre.
\newblock Context-dependent selectivity to natural scenes in the retina.
\newblock \emph{bioRxiv}, 2021.

\bibitem[Hafting et~al.(2005)Hafting, Fyhn, Molden, Moser, and
  Moser]{hafting2005microstructure}
Torkel Hafting, Marianne Fyhn, Sturla Molden, May-Britt Moser, and Edvard~I
  Moser.
\newblock Microstructure of a spatial map in the entorhinal cortex.
\newblock \emph{Nature}, 436\penalty0 (7052):\penalty0 801--806, 2005.

\bibitem[Hamm et~al.(2021)Hamm, Shymkiv, Han, Yang, and
  Yuste]{hamm2021cortical}
Jordan~P Hamm, Yuriy Shymkiv, Shuting Han, Weijian Yang, and Rafael Yuste.
\newblock Cortical ensembles selective for context.
\newblock \emph{Proceedings of the National Academy of Sciences}, 118\penalty0
  (14), 2021.

\bibitem[Hardcastle et~al.(2017)Hardcastle, Maheswaranathan, Ganguli, and
  Giocomo]{hardcastle2017multiplexed}
Kiah Hardcastle, Niru Maheswaranathan, Surya Ganguli, and Lisa~M Giocomo.
\newblock A multiplexed, heterogeneous, and adaptive code for navigation in
  medial entorhinal cortex.
\newblock \emph{Neuron}, 94\penalty0 (2):\penalty0 375--387, 2017.

\bibitem[Harvey et~al.(2012)Harvey, Coen, and Tank]{harvey2012choice}
Christopher~D Harvey, Philip Coen, and David~W Tank.
\newblock Choice-specific sequences in parietal cortex during a
  virtual-navigation decision task.
\newblock \emph{Nature}, 484\penalty0 (7392):\penalty0 62--68, 2012.

\bibitem[Hubel \& Wiesel(1979)Hubel and Wiesel]{hubel1979brain}
D.~H. Hubel and T.~N. Wiesel.
\newblock Brain mechanisms of vision.
\newblock \emph{Scientific American}, 241:\penalty0 130, 1979.

\bibitem[Jacobs et~al.(2013)Jacobs, Weidemann, Miller, Solway, Burke, Wei,
  Suthana, Sperling, Sharan, Fried, et~al.]{jacobs2013direct}
Joshua Jacobs, Christoph~T Weidemann, Jonathan~F Miller, Alec Solway, John~F
  Burke, Xue-Xin Wei, Nanthia Suthana, Michael~R Sperling, Ashwini~D Sharan,
  Itzhak Fried, et~al.
\newblock Direct recordings of grid-like neuronal activity in human spatial
  navigation.
\newblock \emph{Nature neuroscience}, 16\penalty0 (9):\penalty0 1188--1190,
  2013.

\bibitem[Jensen et~al.(2020)Jensen, Kao, Tripodi, and
  Hennequin]{jensen2020manifold}
Kristopher Jensen, Ta-Chu Kao, Marco Tripodi, and Guillaume Hennequin.
\newblock Manifold gplvms for discovering non-euclidean latent structure in
  neural data.
\newblock \emph{Advances in Neural Information Processing Systems},
  33:\penalty0 22580--22592, 2020.

\bibitem[Jensen et~al.(2022)Jensen, Liu, Kao, Lengyel, and
  Hennequin]{jensen2022beyond}
Kristopher~T Jensen, David Liu, Ta-Chu Kao, M{\'a}t{\'e} Lengyel, and Guillaume
  Hennequin.
\newblock Beyond the euclidean brain: inferring non-euclidean latent
  trajectories from spike trains.
\newblock \emph{bioRxiv}, 2022.

\bibitem[Jun et~al.(2017)Jun, Steinmetz, Siegle, Denman, Bauza, Barbarits, Lee,
  Anastassiou, Andrei, Ayd{\i}n, et~al.]{jun2017fully}
James~J Jun, Nicholas~A Steinmetz, Joshua~H Siegle, Daniel~J Denman, Marius
  Bauza, Brian Barbarits, Albert~K Lee, Costas~A Anastassiou, Alexandru Andrei,
  {\c{C}}a{\u{g}}atay Ayd{\i}n, et~al.
\newblock Fully integrated silicon probes for high-density recording of neural
  activity.
\newblock \emph{Nature}, 551\penalty0 (7679):\penalty0 232--236, 2017.

\bibitem[Kanter et~al.(2017)Kanter, Lykken, Avesar, Weible, Dickinson, Dunn,
  Borgesius, Roudi, and Kentros]{kanter2017novel}
Benjamin~R Kanter, Christine~M Lykken, Daniel Avesar, Aldis Weible, Jasmine
  Dickinson, Benjamin Dunn, Nils~Z Borgesius, Yasser Roudi, and Clifford~G
  Kentros.
\newblock A novel mechanism for the grid-to-place cell transformation revealed
  by transgenic depolarization of medial entorhinal cortex layer ii.
\newblock \emph{Neuron}, 93\penalty0 (6):\penalty0 1480--1492, 2017.

\bibitem[Kao et~al.(2015)Kao, Nuyujukian, Ryu, Churchland, Cunningham, and
  Shenoy]{kao2015single}
Jonathan~C Kao, Paul Nuyujukian, Stephen~I Ryu, Mark~M Churchland, John~P
  Cunningham, and Krishna~V Shenoy.
\newblock Single-trial dynamics of motor cortex and their applications to
  brain-machine interfaces.
\newblock \emph{Nature communications}, 6\penalty0 (1):\penalty0 1--12, 2015.

\bibitem[Kaufman et~al.(2014)Kaufman, Churchland, Ryu, and
  Shenoy]{kaufman2014cortical}
Matthew~T Kaufman, Mark~M Churchland, Stephen~I Ryu, and Krishna~V Shenoy.
\newblock Cortical activity in the null space: permitting preparation without
  movement.
\newblock \emph{Nature neuroscience}, 17\penalty0 (3):\penalty0 440--448, 2014.

\bibitem[Kell \& McDermott(2019)Kell and McDermott]{kell2019deep}
Alexander~JE Kell and Josh~H McDermott.
\newblock Deep neural network models of sensory systems: windows onto the role
  of task constraints.
\newblock \emph{Current opinion in neurobiology}, 55:\penalty0 121--132, 2019.

\bibitem[Kell et~al.(2018)Kell, Yamins, Shook, Norman-Haignere, and
  McDermott]{kell2018task}
Alexander~JE Kell, Daniel~LK Yamins, Erica~N Shook, Sam~V Norman-Haignere, and
  Josh~H McDermott.
\newblock A task-optimized neural network replicates human auditory behavior,
  predicts brain responses, and reveals a cortical processing hierarchy.
\newblock \emph{Neuron}, 98\penalty0 (3):\penalty0 630--644, 2018.

\bibitem[Kingma \& Ba(2014)Kingma and Ba]{kingma2014adam}
Diederik~P Kingma and Jimmy Ba.
\newblock Adam: A method for stochastic optimization.
\newblock \emph{arXiv preprint arXiv:1412.6980}, 2014.

\bibitem[Kingma \& Welling(2014)Kingma and Welling]{kingma2013auto}
Diederik~P Kingma and Max Welling.
\newblock Auto-encoding variational bayes.
\newblock In \emph{Proceedings of the International Conference on Learning
  Representations (ICLR)}, 2014.

\bibitem[Klindt et~al.(2017)Klindt, Ecker, Euler, and Bethge]{klindt2017neural}
David Klindt, Alexander~S Ecker, Thomas Euler, and Matthias Bethge.
\newblock Neural system identification for large populations separating
  “what” and “where”.
\newblock \emph{Advances in Neural Information Processing Systems}, 30, 2017.

\bibitem[Klindt et~al.(2021)Klindt, Schott, Sharma, Ustyuzhaninov, Brendel,
  Bethge, and Paiton]{klindt2021towards}
David~A. Klindt, Lukas Schott, Yash Sharma, Ivan Ustyuzhaninov, Wieland
  Brendel, Matthias Bethge, and Dylan Paiton.
\newblock Towards nonlinear disentanglement in natural data with temporal
  sparse coding.
\newblock In \emph{International Conference on Learning Representations}, 2021.
\newblock URL \url{https://openreview.net/forum?id=EbIDjBynYJ8}.

\bibitem[Klukas et~al.(2020)Klukas, Lewis, and Fiete]{klukas2020efficient}
Mirko Klukas, Marcus Lewis, and Ila Fiete.
\newblock Efficient and flexible representation of higher-dimensional cognitive
  variables with grid cells.
\newblock \emph{PLoS computational biology}, 16\penalty0 (4):\penalty0
  e1007796, 2020.

\bibitem[Kriegeskorte \& Wei(2021)Kriegeskorte and Wei]{kriegeskorte2021neural}
Nikolaus Kriegeskorte and Xue-Xin Wei.
\newblock Neural tuning and representational geometry.
\newblock \emph{Nature Reviews Neuroscience}, 22\penalty0 (11):\penalty0
  703--718, 2021.

\bibitem[Kropff et~al.(2015)Kropff, Carmichael, Moser, and
  Moser]{kropff2015speed}
Emilio Kropff, James~E Carmichael, May-Britt Moser, and Edvard~I Moser.
\newblock Speed cells in the medial entorhinal cortex.
\newblock \emph{Nature}, 523\penalty0 (7561):\penalty0 419--424, 2015.

\bibitem[Lawrence(2003)]{lawrence2003gaussian}
Neil Lawrence.
\newblock Gaussian process latent variable models for visualisation of high
  dimensional data.
\newblock \emph{Advances in neural information processing systems}, 16, 2003.

\bibitem[Lewicki(2002)]{lewicki2002efficient}
Michael~S Lewicki.
\newblock Efficient coding of natural sounds.
\newblock \emph{Nature neuroscience}, 5\penalty0 (4):\penalty0 356--363, 2002.

\bibitem[Lieber \& Bensmaia(2019)Lieber and Bensmaia]{lieber2019high}
Justin~D Lieber and Sliman~J Bensmaia.
\newblock High-dimensional representation of texture in somatosensory cortex of
  primates.
\newblock \emph{Proceedings of the National Academy of Sciences}, 116\penalty0
  (8):\penalty0 3268--3277, 2019.

\bibitem[Lopes-dos Santos et~al.(2011)Lopes-dos Santos, Conde-Ocazionez,
  Nicolelis, Ribeiro, and Tort]{lopes2011neuronal}
V{\'\i}tor Lopes-dos Santos, Sergio Conde-Ocazionez, Miguel~AL Nicolelis,
  Sidarta~T Ribeiro, and Adriano~BL Tort.
\newblock Neuronal assembly detection and cell membership specification by
  principal component analysis.
\newblock \emph{PloS one}, 6\penalty0 (6):\penalty0 e20996, 2011.

\bibitem[Lurz et~al.(2021)Lurz, Bashiri, Willeke, Jagadish, Wang, Walker,
  Cadena, Muhammad, Cobos, Tolias, et~al.]{lurz2021generalization}
Konstantin-Klemens Lurz, Mohammad Bashiri, Konstantin Willeke, Akshay~K
  Jagadish, Eric Wang, Edgar~Y Walker, Santiago~A Cadena, Taliah Muhammad,
  Erick Cobos, Andreas~S Tolias, et~al.
\newblock Generalization in data-driven models of primary visual cortex.
\newblock \emph{bioRxiv}, pp.\  2020--10, 2021.

\bibitem[Mante et~al.(2013)Mante, Sussillo, Shenoy, and
  Newsome]{mante2013context}
Valerio Mante, David Sussillo, Krishna~V Shenoy, and William~T Newsome.
\newblock Context-dependent computation by recurrent dynamics in prefrontal
  cortex.
\newblock \emph{Nature}, 503\penalty0 (7474):\penalty0 78--84, 2013.

\bibitem[Mathis et~al.(2012)Mathis, Herz, and Stemmler]{mathis2012optimal}
Alexander Mathis, Andreas~VM Herz, and Martin Stemmler.
\newblock Optimal population codes for space: grid cells outperform place
  cells.
\newblock \emph{Neural computation}, 24\penalty0 (9):\penalty0 2280--2317,
  2012.

\bibitem[McNaughton et~al.(2006)McNaughton, Battaglia, Jensen, Moser, and
  Moser]{mcnaughton2006path}
Bruce~L McNaughton, Francesco~P Battaglia, Ole Jensen, Edvard~I Moser, and
  May-Britt Moser.
\newblock Path integration and the neural basis of the'cognitive map'.
\newblock \emph{Nature Reviews Neuroscience}, 7\penalty0 (8):\penalty0
  663--678, 2006.

\bibitem[Mimica et~al.(2018)Mimica, Dunn, Tombaz, Bojja, and
  Whitlock]{mimica2018efficient}
Bartul Mimica, Benjamin~A Dunn, Tuce Tombaz, VPTNC~Srikanth Bojja, and
  Jonathan~R Whitlock.
\newblock Efficient cortical coding of 3d posture in freely behaving rats.
\newblock \emph{Science}, 362\penalty0 (6414):\penalty0 584--589, 2018.

\bibitem[Nayebi et~al.(2021)Nayebi, Sagastuy-Brena, Bear, Kar, Kubilius,
  Ganguli, Sussillo, DiCarlo, and Yamins]{nayebi2021goal}
Aran Nayebi, Javier Sagastuy-Brena, Daniel~M Bear, Kohitij Kar, Jonas Kubilius,
  Surya Ganguli, David Sussillo, James~J DiCarlo, and Daniel~LK Yamins.
\newblock Goal-driven recurrent neural network models of the ventral visual
  stream.
\newblock \emph{bioRxiv}, 2021.

\bibitem[Pandarinath et~al.(2018)Pandarinath, O’Shea, Collins, Jozefowicz,
  Stavisky, Kao, Trautmann, Kaufman, Ryu, Hochberg,
  et~al.]{pandarinath2018inferring}
Chethan Pandarinath, Daniel~J O’Shea, Jasmine Collins, Rafal Jozefowicz,
  Sergey~D Stavisky, Jonathan~C Kao, Eric~M Trautmann, Matthew~T Kaufman,
  Stephen~I Ryu, Leigh~R Hochberg, et~al.
\newblock Inferring single-trial neural population dynamics using sequential
  auto-encoders.
\newblock \emph{Nature methods}, 15\penalty0 (10):\penalty0 805--815, 2018.

\bibitem[Paszke et~al.(2019)Paszke, Gross, Massa, Lerer, Bradbury, Chanan,
  Killeen, Lin, Gimelshein, Antiga, Desmaison, Kopf, Yang, DeVito, Raison,
  Tejani, Chilamkurthy, Steiner, Fang, Bai, and Chintala]{NEURIPS2019_9015}
Adam Paszke, Sam Gross, Francisco Massa, Adam Lerer, James Bradbury, Gregory
  Chanan, Trevor Killeen, Zeming Lin, Natalia Gimelshein, Luca Antiga, Alban
  Desmaison, Andreas Kopf, Edward Yang, Zachary DeVito, Martin Raison, Alykhan
  Tejani, Sasank Chilamkurthy, Benoit Steiner, Lu~Fang, Junjie Bai, and Soumith
  Chintala.
\newblock Pytorch: An imperative style, high-performance deep learning library.
\newblock In H.~Wallach, H.~Larochelle, A.~Beygelzimer, F.~d\textquotesingle
  Alch\'{e}-Buc, E.~Fox, and R.~Garnett (eds.), \emph{Advances in Neural
  Information Processing Systems 32}, pp.\  8024--8035. Curran Associates,
  Inc., 2019.
\newblock URL
  \url{http://papers.neurips.cc/paper/9015-pytorch-an-imperative-style-high-performance-deep-learning-library.pdf}.

\bibitem[Pei et~al.(2021)Pei, Ye, Zoltowski, Wu, Chowdhury, Sohn, O’Doherty,
  Shenoy, Kaufman, Churchland, Jazayeri, Miller, Pillow, Park, Dyer, and
  Pandarinath]{PeiYe2021NeuralLatents}
Felix Pei, Joel Ye, David~M. Zoltowski, Anqi Wu, Raeed~H. Chowdhury, Hansem
  Sohn, Joseph~E. O’Doherty, Krishna~V. Shenoy, Matthew~T. Kaufman, Mark
  Churchland, Mehrdad Jazayeri, Lee~E. Miller, Jonathan Pillow, Il~Memming
  Park, Eva~L. Dyer, and Chethan Pandarinath.
\newblock Neural latents benchmark '21: Evaluating latent variable models of
  neural population activity.
\newblock In \emph{Advances in Neural Information Processing Systems (NeurIPS),
  Track on Datasets and Benchmarks}, 2021.
\newblock URL \url{https://arxiv.org/abs/2109.04463}.

\bibitem[Peyrache et~al.(2015{\natexlab{a}})Peyrache, Lacroix, Petersen, and
  Buzs{\'a}ki]{peyrache2015internally}
Adrien Peyrache, Marie~M Lacroix, Peter~C Petersen, and Gy{\"o}rgy Buzs{\'a}ki.
\newblock Internally organized mechanisms of the head direction sense.
\newblock \emph{Nature neuroscience}, 18\penalty0 (4):\penalty0 569--575,
  2015{\natexlab{a}}.

\bibitem[Peyrache et~al.(2015{\natexlab{b}})Peyrache, Petersen, and
  Buzs{\'a}ki]{peyrache2015rec}
Adrien Peyrache, Peter~C Petersen, and Gy{\"o}rgy Buzs{\'a}ki.
\newblock Extracellular recordings from multi-site silicon probes in the
  anterior thalamus and subicular formation of freely moving mice.
\newblock CRCNS.org. \url{http://dx.doi.org/10.6080/K0G15XS1},
  2015{\natexlab{b}}.

\bibitem[Qiu et~al.(2020)Qiu, Zhao, Klindt, Kautzky, Szatko, Schaeffel, Rifai,
  Franke, Busse, and Euler]{qiu2020mouse}
Yongrong Qiu, Zhijian Zhao, David Klindt, Magdalena Kautzky, Klaudia~P Szatko,
  Frank Schaeffel, Katharina Rifai, Katrin Franke, Laura Busse, and Thomas
  Euler.
\newblock Mouse retinal specializations reflect knowledge of natural
  environment statistics.
\newblock \emph{bioRxiv}, 2020.

\bibitem[Rezende et~al.(2014)Rezende, Mohamed, and
  Wierstra]{rezende2014stochastic}
Danilo~Jimenez Rezende, Shakir Mohamed, and Daan Wierstra.
\newblock Stochastic backpropagation and approximate inference in deep
  generative models.
\newblock In \emph{International conference on machine learning}, pp.\
  1278--1286. PMLR, 2014.

\bibitem[Rubin et~al.(2019)Rubin, Sheintuch, Brande-Eilat, Pinchasof, Rechavi,
  Geva, and Ziv]{rubin2019revealing}
Alon Rubin, Liron Sheintuch, Noa Brande-Eilat, Or~Pinchasof, Yoav Rechavi,
  Nitzan Geva, and Yaniv Ziv.
\newblock Revealing neural correlates of behavior without behavioral
  measurements.
\newblock \emph{Nature communications}, 10\penalty0 (1):\penalty0 1--14, 2019.

\bibitem[Rule et~al.(2019)Rule, O’Leary, and Harvey]{rule2019causes}
Michael~E Rule, Timothy O’Leary, and Christopher~D Harvey.
\newblock Causes and consequences of representational drift.
\newblock \emph{Current opinion in neurobiology}, 58:\penalty0 141--147, 2019.

\bibitem[Rybakken et~al.(2019)Rybakken, Baas, and Dunn]{rybakken2019decoding}
Erik Rybakken, Nils Baas, and Benjamin Dunn.
\newblock Decoding of neural data using cohomological feature extraction.
\newblock \emph{Neural computation}, 31\penalty0 (1):\penalty0 68--93, 2019.

\bibitem[Sadtler et~al.(2014)Sadtler, Quick, Golub, Chase, Ryu, Tyler-Kabara,
  Yu, and Batista]{sadtler2014neural}
Patrick~T Sadtler, Kristin~M Quick, Matthew~D Golub, Steven~M Chase, Stephen~I
  Ryu, Elizabeth~C Tyler-Kabara, Byron~M Yu, and Aaron~P Batista.
\newblock Neural constraints on learning.
\newblock \emph{Nature}, 512\penalty0 (7515):\penalty0 423--426, 2014.

\bibitem[Safarani et~al.(2021)Safarani, Nix, Willeke, Cadena, Restivo,
  Denfield, Tolias, and Sinz]{safarani2021towards}
Shahd Safarani, Arne Nix, Konstantin Willeke, Santiago Cadena, Kelli Restivo,
  George Denfield, Andreas Tolias, and Fabian Sinz.
\newblock Towards robust vision by multi-task learning on monkey visual cortex.
\newblock \emph{Advances in Neural Information Processing Systems}, 34, 2021.

\bibitem[Sargolini et~al.(2006)Sargolini, Fyhn, Hafting, McNaughton, Witter,
  Moser, and Moser]{sargolini2006conjunctive}
Francesca Sargolini, Marianne Fyhn, Torkel Hafting, Bruce~L McNaughton, Menno~P
  Witter, May-Britt Moser, and Edvard~I Moser.
\newblock Conjunctive representation of position, direction, and velocity in
  entorhinal cortex.
\newblock \emph{Science}, 312\penalty0 (5774):\penalty0 758--762, 2006.

\bibitem[Schneidman et~al.(2006)Schneidman, Berry, Segev, and
  Bialek]{schneidman2006weak}
Elad Schneidman, Michael~J Berry, Ronen Segev, and William Bialek.
\newblock Weak pairwise correlations imply strongly correlated network states
  in a neural population.
\newblock \emph{Nature}, 440\penalty0 (7087):\penalty0 1007--1012, 2006.

\bibitem[Schott et~al.(2018)Schott, Rauber, Bethge, and
  Brendel]{schott2018towards}
Lukas Schott, Jonas Rauber, Matthias Bethge, and Wieland Brendel.
\newblock Towards the first adversarially robust neural network model on mnist.
\newblock \emph{arXiv preprint arXiv:1805.09190}, 2018.

\bibitem[Seeliger et~al.(2021)Seeliger, Ambrogioni,
  G{\"u}{\c{c}}l{\"u}t{\"u}rk, van~den Bulk, G{\"u}{\c{c}}l{\"u}, and van
  Gerven]{seeliger2021end}
Katja Seeliger, Luca Ambrogioni, Ya{\u{g}}mur G{\"u}{\c{c}}l{\"u}t{\"u}rk,
  Leonieke~M van~den Bulk, Umut G{\"u}{\c{c}}l{\"u}, and MAJ van Gerven.
\newblock End-to-end neural system identification with neural information flow.
\newblock \emph{PLOS Computational Biology}, 17\penalty0 (2):\penalty0
  e1008558, 2021.

\bibitem[Shah et~al.(2022)Shah, Brackbill, Samarakoon, Rhoades, Kling, Sher,
  Litke, Singer, Shlens, and Chichilnisky]{shah2022individual}
Nishal~P Shah, Nora Brackbill, Ryan Samarakoon, Colleen Rhoades, Alexandra
  Kling, Alexander Sher, Alan Litke, Yoram Singer, Jonathon Shlens, and
  EJ~Chichilnisky.
\newblock Individual variability of neural computations in the primate retina.
\newblock \emph{Neuron}, 110\penalty0 (4):\penalty0 698--708, 2022.

\bibitem[Shenoy et~al.(2013)Shenoy, Sahani, and Churchland]{shenoy2013cortical}
Krishna~V Shenoy, Maneesh Sahani, and Mark~M Churchland.
\newblock Cortical control of arm movements: a dynamical systems perspective.
\newblock \emph{Annual review of neuroscience}, 36:\penalty0 337--359, 2013.

\bibitem[Singh et~al.(2008)Singh, Memoli, Ishkhanov, Sapiro, Carlsson, and
  Ringach]{singh2008topological}
Gurjeet Singh, Facundo Memoli, Tigran Ishkhanov, Guillermo Sapiro, Gunnar
  Carlsson, and Dario~L Ringach.
\newblock Topological analysis of population activity in visual cortex.
\newblock \emph{Journal of vision}, 8\penalty0 (8):\penalty0 11--11, 2008.

\bibitem[Sinz et~al.(2018)Sinz, Ecker, Fahey, Walker, Cobos, Froudarakis,
  Yatsenko, Pitkow, Reimer, and Tolias]{sinz2018stimulus}
Fabian Sinz, Alexander~S Ecker, Paul Fahey, Edgar Walker, Erick Cobos,
  Emmanouil Froudarakis, Dimitri Yatsenko, Zachary Pitkow, Jacob Reimer, and
  Andreas Tolias.
\newblock Stimulus domain transfer in recurrent models for large scale cortical
  population prediction on video.
\newblock \emph{Advances in neural information processing systems}, 31, 2018.

\bibitem[Skaggs et~al.(1992)Skaggs, Mcnaughton, and
  Gothard]{skaggs1992information}
William Skaggs, Bruce Mcnaughton, and Katalin Gothard.
\newblock An information-theoretic approach to deciphering the hippocampal
  code.
\newblock \emph{Advances in neural information processing systems}, 5, 1992.

\bibitem[Solstad et~al.(2006)Solstad, Moser, and Einevoll]{solstad2006grid}
Trygve Solstad, Edvard~I Moser, and Gaute~T Einevoll.
\newblock From grid cells to place cells: a mathematical model.
\newblock \emph{Hippocampus}, 16\penalty0 (12):\penalty0 1026--1031, 2006.

\bibitem[Sreenivasan \& Fiete(2011)Sreenivasan and Fiete]{sreenivasan2011grid}
Sameet Sreenivasan and Ila Fiete.
\newblock Grid cells generate an analog error-correcting code for singularly
  precise neural computation.
\newblock \emph{Nature neuroscience}, 14\penalty0 (10):\penalty0 1330--1337,
  2011.

\bibitem[Stensola et~al.(2012)Stensola, Stensola, Solstad, Fr{\o}land, Moser,
  and Moser]{stensola2012entorhinal}
Hanne Stensola, Tor Stensola, Trygve Solstad, Kristian Fr{\o}land, May-Britt
  Moser, and Edvard~I Moser.
\newblock The entorhinal grid map is discretized.
\newblock \emph{Nature}, 492\penalty0 (7427):\penalty0 72--78, 2012.

\bibitem[Stevenson \& Kording(2011)Stevenson and
  Kording]{stevenson2011advances}
Ian~H Stevenson and Konrad~P Kording.
\newblock How advances in neural recording affect data analysis.
\newblock \emph{Nature neuroscience}, 14\penalty0 (2):\penalty0 139--142, 2011.

\bibitem[Stokes et~al.(2013)Stokes, Kusunoki, Sigala, Nili, Gaffan, and
  Duncan]{stokes2013dynamic}
Mark~G Stokes, Makoto Kusunoki, Natasha Sigala, Hamed Nili, David Gaffan, and
  John Duncan.
\newblock Dynamic coding for cognitive control in prefrontal cortex.
\newblock \emph{Neuron}, 78\penalty0 (2):\penalty0 364--375, 2013.

\bibitem[Stringer et~al.(2019)Stringer, Pachitariu, Steinmetz, Carandini, and
  Harris]{stringer2019high}
Carsen Stringer, Marius Pachitariu, Nicholas Steinmetz, Matteo Carandini, and
  Kenneth~D Harris.
\newblock High-dimensional geometry of population responses in visual cortex.
\newblock \emph{Nature}, 571\penalty0 (7765):\penalty0 361--365, 2019.

\bibitem[Taube et~al.(1990)Taube, Muller, and Ranck]{taube1990head}
Jeffrey~S Taube, Robert~U Muller, and James~B Ranck.
\newblock Head-direction cells recorded from the postsubiculum in freely moving
  rats. i. description and quantitative analysis.
\newblock \emph{Journal of Neuroscience}, 10\penalty0 (2):\penalty0 420--435,
  1990.

\bibitem[Tukker et~al.(2022)Tukker, Beed, Brecht, Kempter, Moser, and
  Schmitz]{tukker2022microcircuits}
John~J Tukker, Prateep Beed, Michael Brecht, Richard Kempter, Edvard~I Moser,
  and Dietmar Schmitz.
\newblock Microcircuits for spatial coding in the medial entorhinal cortex.
\newblock \emph{Physiological reviews}, 102\penalty0 (2):\penalty0 653--688,
  2022.

\bibitem[Ustyuzhaninov et~al.(2019)Ustyuzhaninov, Cadena, Froudarakis, Fahey,
  Walker, Cobos, Reimer, Sinz, Tolias, Bethge,
  et~al.]{ustyuzhaninov2019rotation}
Ivan Ustyuzhaninov, Santiago~A Cadena, Emmanouil Froudarakis, Paul~G Fahey,
  Edgar~Y Walker, Erick Cobos, Jacob Reimer, Fabian~H Sinz, Andreas~S Tolias,
  Matthias Bethge, et~al.
\newblock Rotation-invariant clustering of neuronal responses in primary visual
  cortex.
\newblock In \emph{International Conference on Learning Representations}, 2019.

\bibitem[Ustyuzhaninov et~al.(2022)Ustyuzhaninov, Burg, Cadena, Fu, Muhammad,
  Ponder, Froudarakis, Ding, Bethge, Tolias, et~al.]{ustyuzhaninov2022digital}
Ivan Ustyuzhaninov, Max~F Burg, Santiago~A Cadena, Jiakun Fu, Taliah Muhammad,
  Kayla Ponder, Emmanouil Froudarakis, Zhiwei Ding, Matthias Bethge, Andreas
  Tolias, et~al.
\newblock Digital twin reveals combinatorial code of non-linear computations in
  the mouse primary visual cortex.
\newblock \emph{bioRxiv}, 2022.

\bibitem[Walker et~al.(2018)Walker, Sinz, Froudarakis, Fahey, Muhammad, Ecker,
  Cobos, Reimer, Pitkow, and Tolias]{walker2018inception}
Edgar~Y Walker, Fabian~H Sinz, Emmanouil Froudarakis, Paul~G Fahey, Taliah
  Muhammad, Alexander~S Ecker, Erick Cobos, Jacob Reimer, Xaq Pitkow, and
  Andreas~S Tolias.
\newblock Inception in visual cortex: in vivo-silico loops reveal most exciting
  images.
\newblock \emph{bioRxiv}, pp.\  506956, 2018.

\bibitem[W{\"a}ssle \& Riemann(1978)W{\"a}ssle and Riemann]{wassle1978mosaic}
H~W{\"a}ssle and HJ~Riemann.
\newblock The mosaic of nerve cells in the mammalian retina.
\newblock \emph{Proceedings of the Royal Society of London. Series B.
  Biological Sciences}, 200\penalty0 (1141):\penalty0 441--461, 1978.

\bibitem[Wei et~al.(2015)Wei, Prentice, and Balasubramanian]{wei2015principle}
Xue-Xin Wei, Jason Prentice, and Vijay Balasubramanian.
\newblock A principle of economy predicts the functional architecture of grid
  cells.
\newblock \emph{Elife}, 4:\penalty0 e08362, 2015.

\bibitem[Wu et~al.(2017)Wu, Roy, Keeley, and Pillow]{wu2017gaussian}
Anqi Wu, Nicholas~A Roy, Stephen Keeley, and Jonathan~W Pillow.
\newblock Gaussian process based nonlinear latent structure discovery in
  multivariate spike train data.
\newblock \emph{Advances in neural information processing systems}, 30, 2017.

\bibitem[Wu et~al.(2018)Wu, Pashkovski, Datta, and Pillow]{wu2018learning}
Anqi Wu, Stan Pashkovski, Sandeep~R Datta, and Jonathan~W Pillow.
\newblock Learning a latent manifold of odor representations from neural
  responses in piriform cortex.
\newblock \emph{Advances in Neural Information Processing Systems}, 31, 2018.

\bibitem[Yu et~al.(2008)Yu, Cunningham, Santhanam, Ryu, Shenoy, and
  Sahani]{yu2008gaussian}
Byron~M Yu, John~P Cunningham, Gopal Santhanam, Stephen Ryu, Krishna~V Shenoy,
  and Maneesh Sahani.
\newblock Gaussian-process factor analysis for low-dimensional single-trial
  analysis of neural population activity.
\newblock \emph{Advances in neural information processing systems}, 21, 2008.

\bibitem[Zhao et~al.(2020)Zhao, Klindt, Maia~Chagas, Szatko, Rogerson, Protti,
  Behrens, Dalkara, Schubert, Bethge, et~al.]{zhao2020temporal}
Zhijian Zhao, David~A Klindt, Andr{\'e} Maia~Chagas, Klaudia~P Szatko, Luke
  Rogerson, Dario~A Protti, Christian Behrens, Deniz Dalkara, Timm Schubert,
  Matthias Bethge, et~al.
\newblock The temporal structure of the inner retina at a single glance.
\newblock \emph{Scientific reports}, 10\penalty0 (1):\penalty0 1--17, 2020.

\bibitem[Zhou et~al.(2019)Zhou, Barnes, Lu, Yang, and Li]{zhou2019continuity}
Yi~Zhou, Connelly Barnes, Jingwan Lu, Jimei Yang, and Hao Li.
\newblock On the continuity of rotation representations in neural networks.
\newblock In \emph{Proceedings of the IEEE/CVF Conference on Computer Vision
  and Pattern Recognition}, pp.\  5745--5753, 2019.

\bibitem[Zhuang et~al.(2021)Zhuang, Yan, Nayebi, Schrimpf, Frank, DiCarlo, and
  Yamins]{zhuang2021unsupervised}
Chengxu Zhuang, Siming Yan, Aran Nayebi, Martin Schrimpf, Michael~C Frank,
  James~J DiCarlo, and Daniel~LK Yamins.
\newblock Unsupervised neural network models of the ventral visual stream.
\newblock \emph{Proceedings of the National Academy of Sciences}, 118\penalty0
  (3), 2021.

\end{thebibliography}
\bibliographystyle{iclr2023_conference}

\newpage
\appendix
\section*{Appendix}

\section{Model details}\label{section:appendix_model_details}

All models are trained with the Adam optimizer \citep{kingma2014adam}, a learning rate of $0.001$, temporal chunks of size $128$ ($64$ if data length is smaller than $128$) and a batch size of $1$.
Moreover, training is concluded after the training objective has not improved for $5$ steps ($10$ for grid cell data).
All models are implemented in Pytorch \citep{NEURIPS2019_9015}.
The architecture of the encoder is as follows: a 1D grouped convolution with kernel size $9$; followed by two layers of 1D convolutions with kernel size $1$ (i.e., effectively just MLPs at each time point) and 64 filters ($128$ for grid cell data), followed by a linear projection onto the dimensionality of the mean and variance parameters required by the variational posterior.
The decoder follows the feature sharing tuning curve structure as presented in the paper.
All experiments where executed with specified random seeds for data generation, model initialisation and training batch selection to maintain complete reproducibilty.

\subsection{A fast approximation to the objective function}
Next, we present our simplification of the theoretical setup from \citep{davidson2018hyperspherical}, concerning the use of von Mises distributions when inferring latent variables on circular manifolds.
First, we note that as the variance $\sigma^2$ of a Gaussian and the inverse concentration $1/\kappa$ of a von Mises approach $0$, they both become delta functions.
Moreover, they each define the maximum entropy distribution for a fixed variance (concentration) in their respective domains.
Hence, the von Mises is also referred to as the circular normal distribution.

We therefore explored the model behavior when replacing the KL term in Eq. (\ref{eq:elbo}) with a simpler expression, akin to the standard VAE KL, which simply pushes the variance of the variational encoder towards $1$ (i.e., standard normal), but without constraints on the mean (uniform over the latent space, as the full von Mises prior).
Reparameterisation and sampling was then performed by drawing from that Gaussian and adding the scaled perturbation (same as in standard VAEs) to the angle that the variational posterior returned as mean estimate.

In general, performing variational inference in non-Euclidean spaces is challenging \citep{falorsi2019reparameterizing}, which has warranted the use of specialized inference procedures in previous latent variable models on such manifolds \citep{davidson2018hyperspherical, falorsi2019reparameterizing, jensen2020manifold, jensen2022beyond}.
In this work, we follow the `ReLie' approach outlined by \citet{falorsi2019reparameterizing} for general Lie groups, which involves defining a parameterizable (in our case Gaussian) distribution on the tangent space of the group and projecting it onto the manifold using the so-called `exponential map' of the group.
This approach is particularly simple for the toroidal manifolds we consider, where the projection step can be achieved by a simple modulo $2\pi$ operation \citep{jensen2020manifold}.
This leaves the challenge of computing the KL term in Eq. \ref{eq:elbo}, which is given by
\begin{equation}
    \label{eq:KL_app}
    \text{KL}[q(z) | p(z)] = E_{q(z)} \left [ \log q(z) - \log p(z) \right ].
\end{equation}

As a prior $p(z)$, we use a unit Gaussian projected onto the circle, with a mean matched to the posterior mean.
Since the KL divergence in Eq. \ref{eq:KL_app} is invariant to a rotation around the circle \citep{falorsi2019reparameterizing}, we can compute it simply as the KL divergence between two wrapped Gaussian distributions centered at the origin with variances $\sigma^2_q$ and $\sigma^2_p = 1$ respectively.
In general, the projected densities needed in Eq. \ref{eq:KL_app} are given by
\begin{equation}
    \label{eq:relie}
    q_\theta(z) = \sum_{x \in \mathbb{R} \: : \: \exp_G{(x)} = z}{ r_\theta(x) |J(x)|^{-1} },
\end{equation}
as derived in \citet{falorsi2019reparameterizing}. Here, $\exp_G$ is the exponential map, $J(x)$ is the Jacobian at $x$ in the tangent space, and $r_\theta$ is the parameterized reference distribution on the tangent space.

For the circle, the Jacobian is simply given by the identity $J(x) = 1 \, \forall_x$.
With a zero-centered Gaussian reference distribution, we can therefore rewrite Eq. \ref{eq:relie} as 
\begin{equation}
    q_\theta(z) = \sum_{k \in \mathbb{Z}}{ \mathcal{N}(z + 2 k \pi; \mu = 0, \sigma_q^2) },
\end{equation}
and similarly for the prior $p(z)$ \citep{jensen2020manifold}.
To approximate this infinite sum, it is necessary to truncate it after a finite number of elements \citep{falorsi2019reparameterizing, jensen2020manifold, jensen2022beyond}.
In the case where $\sigma_q \ll 1$, it is sufficient to use a single term since there is negligible probability mass outside the area of injectivity ($[\pi, \pi]$).
This is in fact the regime we are operating in, since less than $0.2\%$ of the prior probability mass falls outside the interval $[-\pi, \pi]$, and the posterior variance will in general be smaller than the prior variance.

Additionally, while the integral in Eq. \ref{eq:KL_app} covers the domain $[-\pi, \pi]$, we can approximate it with an integral from $-\infty$ to $\infty$ since the posterior probability mass outside the area of injectivity is negligible, $\int_{x = \pi}^\infty \mathcal{N}(x; 0, \sigma_q^2) dx \approx 0$.
These approximations together imply that we can simply approximate the KL term in Eq. \ref{eq:KL_app} as the KL between our prior and posterior Gaussians \emph{directly in the tangent space}, and we verified numerically that this approximation is excellent for $\sigma_q^2 \leq 1$.
We thus obtain a simple analytical expression for the KL:
\begin{align}
    \text{KL}[q(z) | p(z)] &\approx \text{KL} \left [ \mathcal{N}(z; 0, \sigma_q^2)|  \mathcal{N}(z; 0, 1) \right ] \\
    &= \frac{1}{2} \left ( \sigma_q^2 -1 - \log \sigma_q^2 \right ).
\end{align}
Note that this is only possible because our prior is \emph{localized}, in contrast to previous work which often uses a more diffuse or even uniform prior distribution \citep{falorsi2019reparameterizing, jensen2020manifold}, and because we consider circular latent spaces where the Jacobian is $1$.
Since our prior and posterior both factorize across dimensions, these considerations also generalize to higher dimensional spaces, where the multivariate KL divergence is simply the sum across dimensions of the corresponding univariate KL divergences.

Finally, to verify that these approximations do not affect our results or conclusions, we repeated several key analyses with the variational inference framework described in \citet{davidson2018hyperspherical}. The simplified setup trained about $5 \times$ faster, while behaving nearly identically on all test cases that we evaluated, hence, all experiments in the paper are performed with the simplified objective.
We note that this simplified procedure may be of interest to those who might need to iterate over models without access to a large compute cluster.


\section{Data Generation for simulations}\label{section:data_generation_for_sims}

When conducting the experiments that showcase the scaling of various models with respect to the amount of data (Sec. \ref{sec:feature_sharing_sims}), we generate the spikes according to Gaussian-like tuning curves. More specifically, we assume a common shape for each curve (location assigned randomly on $\mathbb{S}^1$), with a tuning width of $1.2$ radians. Peak rates are set to $0.5$ spikes per bin, the background rate to $0.005$, with a signal-to-noise ratio of $1$ for the Poisson noise. The latent variable is simulated according to a Gaussian process with kernel standard deviation equal to $5.0$, and with kernel scale $50.0$. It is then projected onto $\mathbb{S}^1$ via the modulo operation. The ensemble detection experiments in Sec. \ref{sec:ensemble_detection} use the same data generation, only with multiple ($2$ or $3$) ensembles, with distinct latent varibles for each ensemble.

\section{Ensemble detection comparisons}\label{section:ensemble_detection_comparisons_explained}
For \emph{raw-pca}, we performed k-means clustering on dimensionally reduced neural activity over time \citep[loosely inspired by][]{lopes2011neuronal,baden2016functional,hamm2021cortical}.
Specifically, we took the projection of the spike matrix onto its first $8$ (cross-validated) principal components.
We then computed k-means clustering on this reduced matrix with the standard \href{https://scikit-learn.org/stable/modules/generated/sklearn.cluster.KMeans.html}{scikit-learn} implementation, using a fixed number of $2$ ($3$, depending on the setting) clusters, $100$ initializations and $1000$ maximum iterations.

For \emph{cov-pca}, we performed clustering on the neural covariance matrix \citep[loosely inspired by][]{carrillo2015endogenous}.
Specifically, we computed the covariance across neurons from the spike matrix.
Again, we computed the first $8$ (cross-validated) principal components of the absolute value of this covariance matrix.
The absolute value was taken to preserve any negative or positive dependencies across neurons, as one would presumably observe within an ensemble --- removing the modulus operator drastically reduced ensemble detection accuracy.
From the projection onto the principal components we proceeded again as above.
We also performed a clustering of the inverse of the covariance matrix, as a first order approximation to an Ising model \citep{schneidman2006weak}, but this approach did not yield results above chance level, hence it was not included in the figures.

Finally, for the \emph{supervised} method we performed clustering on the mutual information score (see \href{https://scikit-learn.org/stable/modules/generated/sklearn.feature_selection.mutual_info_regression.html}{scikit-learn} for implementation) between each neuron's activity, and the true latent variables, as input.
Specifically, the features $y$ that were used as input to k-means (same settings as above) were $y_{ij} \sim I[x_i; z_j]$ for every neuron $i$ and every \emph{ground truth} latent dimension $j$.
This presents an upper bound to the achievable performance, since it assumes \emph{supervised} knowledge of the true latent variable values.

\section{The Evidence Lower Bound (ELBO)}\label{section:ELBO}

Here we show the derivation of the ELBO, which is maximized as part of the objective function during the model training procedure

\begin{equation}\label{eq:elbo}
    \begin{split}
        \log p(x) &= \log \int p(x|z) p(z) dz
        = \log \int p(x|z) p(z) \frac{q(z|x)}{q(z|x)} dz \\
        &= \log E_{z \sim q(z|x)}\left[p(x|z) \frac{p(z)}{q(z|x)}\right]
        \overset{\text{Jensen's inequality}}{\geq} E_{z \sim q(z|x)}\left[\log p(x|z) \frac{p(z)}{q(z|x)}\right] \\
        &= E_{z \sim q(z|x)} \underbrace{\left[\log p(x|z)\right]}_{=: \mathcal{L}_{\text{rec}}(x, z)} - \underbrace{E_{z \sim q(z|x)}\left[\log q(z|x) - \log p(z)\right]}_{\text{= KL(q(z|x)|p(z))}} =: \text{ELBO}(q).\\
    \end{split}
\end{equation}

\subsection{Temporal transition priors}\label{section:temporal_transition_priors}

All experiments in the article are conducted using the objective function from Eq. \ref{eq:elbo}. Extending this with additional terms however, e.g. by including a Laplacian temporal transition prior \citep{klindt2021towards} or a Gaussian transition prior (i.e. an $L_1$ or $L_2$ penalty across latent time steps, respectively) is relatively simple, and both the $L_1$ and $L_2$ penalties are included as optional regularization terms in our model. While exhaustive experiments have not been performed regarding these priors, results seem to indicate that the inclusion of these priors do little to improve the model performance in the settings we have studied in the article, which is in agreement with results in Appendix Section \ref{section: ablation_conv_filter} and \ref{section: hyperparams} regarding the convolutional kernel size.

\section{Ablation study --- convolutional filter }\label{section: ablation_conv_filter}

As mentioned in Sec. \ref{sec:feature_sharing_sims}, we conduct a similar model performance comparison between the faeLVMs and mGPLVM, while reducing the kernel size of the convolutional layer for the faeLVMs to $1$ (effectively allowing the encoder network to let the variational posterior only depend on the instantaneous neural population activity). This is done to reduce the potential gap in information content available between a convolutional and a non-convolutional model. Note, however, that this applies only to the variational posterior. At inference time, both models infer the best latent at each point in time separately. Experimental settings are precisely the same as those in Sec. \ref{sec:feature_sharing_sims}, other than the size of the convolutional kernel.

\begin{figure}[h]
  \centering
  \includegraphics[width=1\textwidth]{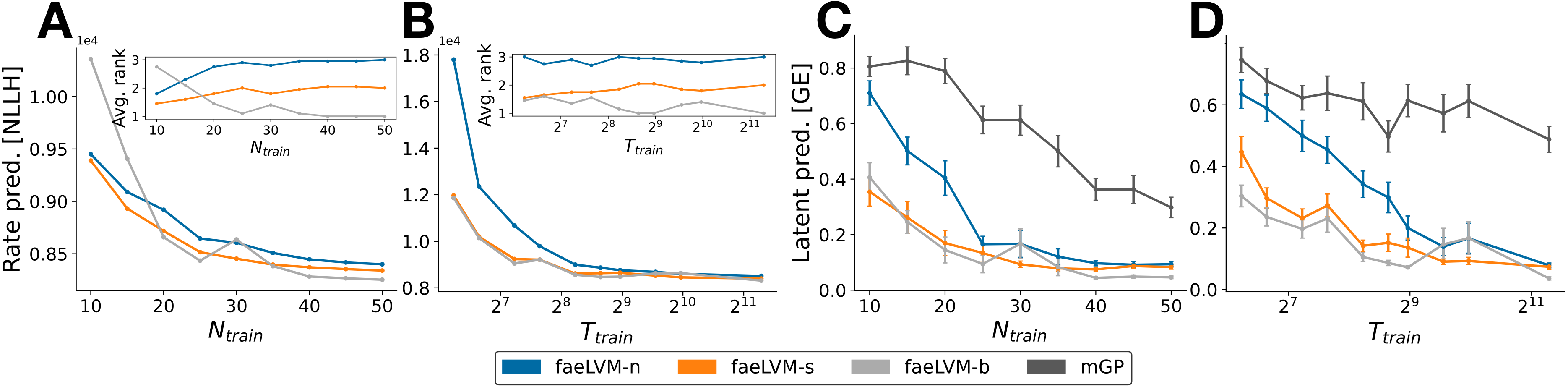}
  \caption{\textbf{Model performance comparison, with convolutional filter size $\mathbf{1}$}. The models faeLVM-{n,s,b} are described in the main text, mGP is the model from \citet{jensen2020manifold}. \textbf{A}, Mean NLLH (negative log-likelihood, lower is better) of predicted test neuron rate, as a function of amount of training neurons. As the performance is averaged over $20$ seeds of varying data, error bars would also reflect the variability in NLLH across data, resulting in inappropriate visualization. Hence, we report the mean rank of $20$ seeds for each model at each condition, in an effort to more appropriately capture the variability of the models. \textbf{B}, Same as in A, now as a function of training time points. \textbf{C}, Mean GE (geodesic error, lower is better) between true latent and inferred latent, on test data, with corresponding SEM error bars, as a function of number of training neurons. \textbf{D}, Same as in C, now as a function of amount of training time points.}
  \label{Kernel_1_scaling}
\end{figure}

One of the main observations from Fig. \ref{Kernel_1_scaling}C-D is that the faeLVMs still outperform mGPLVM, even without contributions from the convolutional kernel. In fact, faeLVM-s performs much better now, compared with the results from Fig. \ref{Scaling_in_T_and_N}, achieving results that rivals those of faeLVM-b (Fig. \ref{Kernel_1_scaling}A-B)

\section{Model extension --- calcium responses and visual data simulation}\label{section:calcium_visual_example}

\begin{figure}
\vspace{10pt}
  \centering
  \includegraphics[width=1\textwidth]{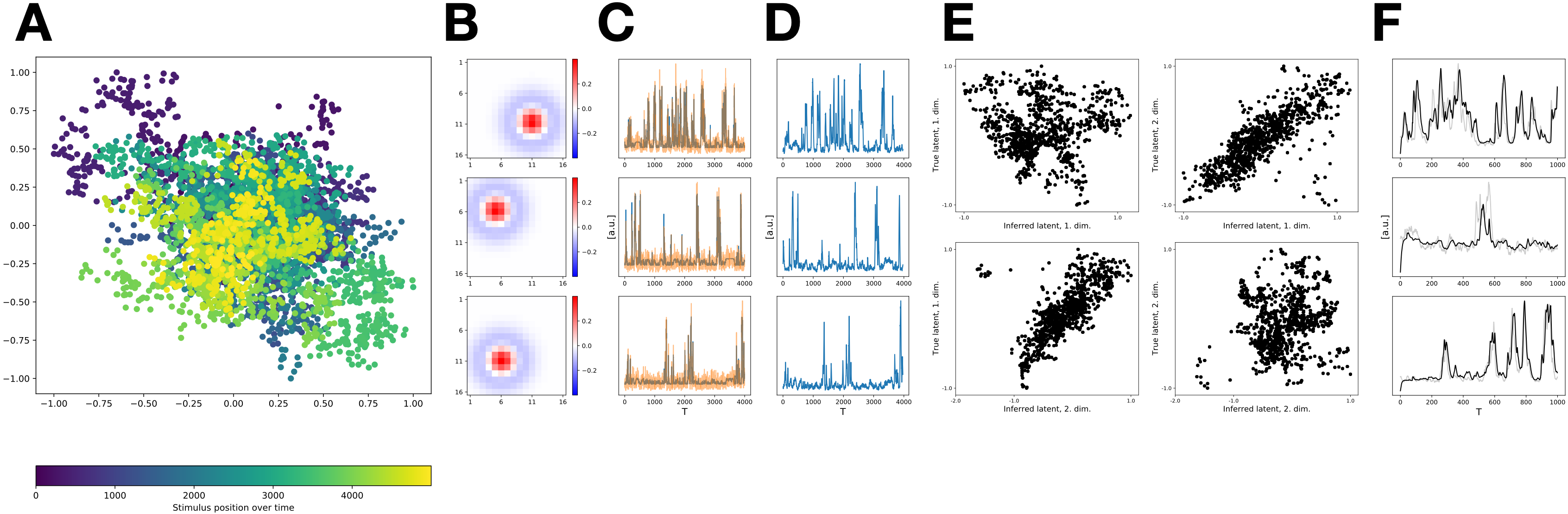}
  \caption{\textbf{Visual data, toy example}. \textbf{A}, Scatter plot of trajectory of simulated visual stimulus, colored according to its temporal evolution. \textbf{B}, Receptive fields for three training neurons. \textbf{C}, Same three training neurons, showing firing rate in shaded dark blue and the corresponding Poisson spikes in orange. \textbf{D}, Neural responses after passing the neural activity from C through a calcium kernel. \textbf{E}, Scatter plot of true latent trajectory against inferred latent trajectory (variational mean). \textbf{F} True Ca2+ neural responses (grey) plotted against predicted neural responses for three test neurons (black).
}
  \label{Visual_calcium_experiment}
\end{figure}

In an effort to improve on the applicability of faeLVM, and showcase its relevance regarding applications to different brain regions, different latent topologies and different input data, we include a toy example on visual data, where we modeled a moving dot stimulus (Fig. \ref{Visual_calcium_experiment}A) and a population response of retinal ganglion cells with center surround receptive fields and Poisson spiking (Fig. \ref{Visual_calcium_experiment}B-C). Each neuron's spiking response was then convolved with a double exponential calcium fluorescence kernel (Fig. \ref{Visual_calcium_experiment}D). We further extended faeLVM by including a Gaussian likelihood as a modeling option, as well as an option to infer latent trajectories not restricted to circular and toroidal latent manifolds.

Results can be seen in Fig. \ref{Visual_calcium_experiment}E-F, where the model is able to infer the correct two-dimensional latent variable, as well as accurately predict calcium responses of test neurons on test data. Although this experiment is less comprehensive, being a toy example, it still gives a clear indication that our model is also applicable to both calcium traces and non-grid and head direction cells.

\section{All grid cell rate maps}\label{section: all_grid_cell_maps}

\begin{figure}
  \centering
  \includegraphics[width=1\textwidth]{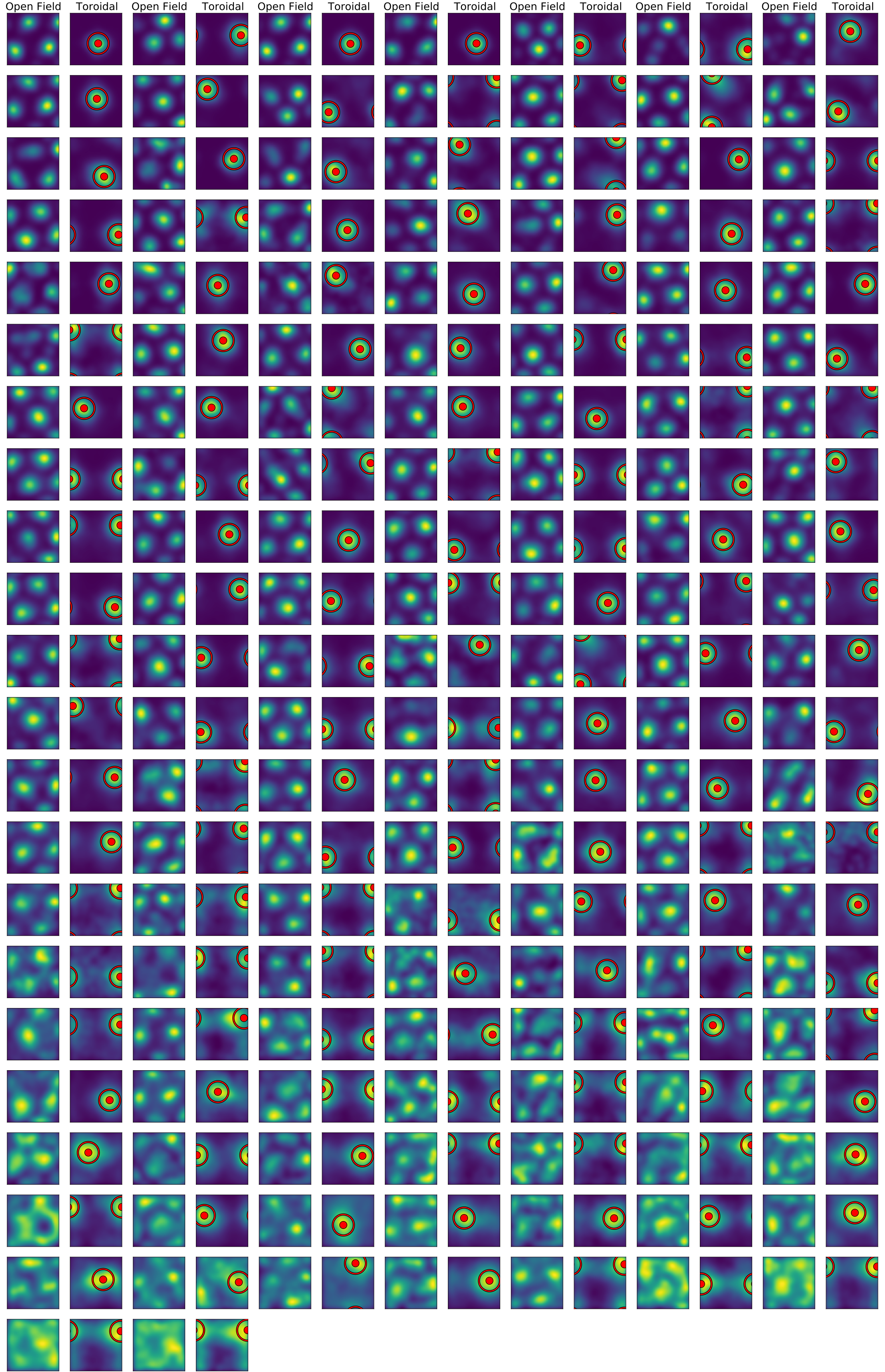}
  \caption{\textbf{All rate maps, first grid cell module}.}
  \label{All_rfs_mod2}
\end{figure}

\begin{figure}
  \centering
  \includegraphics[width=1\textwidth]{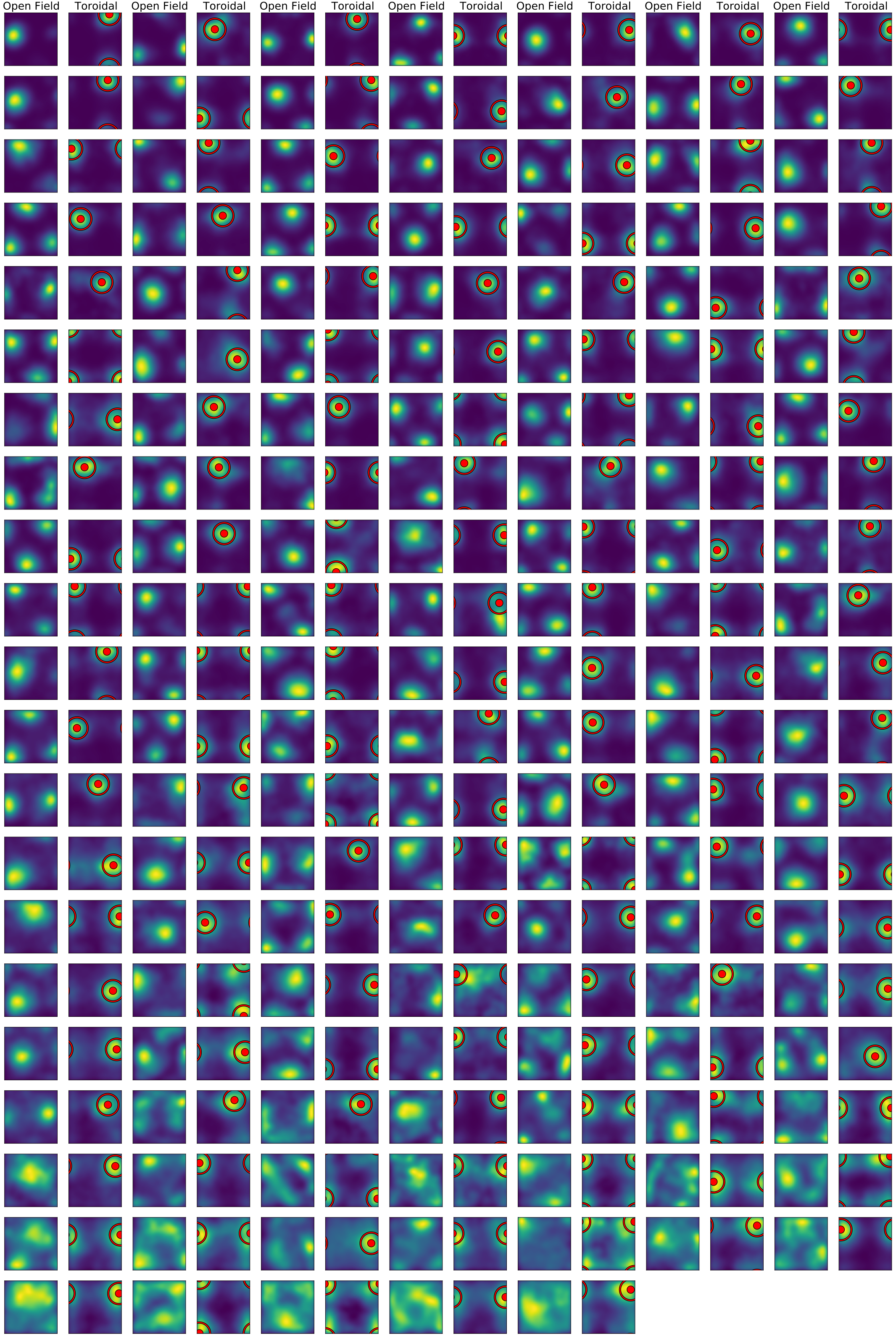}
  \caption{\textbf{All rate maps, second grid cell module}.}
  \label{All_rfs_mod3}
\end{figure}

In this section, we showcase rate maps for all neurons in each of the two ensembles ($149$ and $145$ neurons in the first and second grid cell module respectively). The data used is the same as discussed in Sec. \ref{sec:grid cells data}, pre-processed in the same way, and we used the same model to produce latent decodings, receptive fields and ensembles.

Fig. \ref{All_rfs_mod2} and Fig. \ref{All_rfs_mod3} show the rate maps (using $50 \times 50$ spatial bins, smoothed for visualization purposes in the same manner as \citet{gardner2022toroidal}, with a Gaussian kernel of smoothing width $2.75$), both as function of spatial position and inferred toroidal coordinates. For the rate maps in torodial coordinates, we also include the inferred receptive field center and width. Neural rate maps are shown in decreasing spatial information content \citep{skaggs1992information,gardner2022toroidal}, going from left to right, top to bottom.

As mentioned in Sec. \ref{sec:grid cells data}, we clearly see the more ambiguous rate maps at the bottom of these two figures (in particular, the three-four bottom-most rows). Here, the spatial resolution is not as distinct as for the more informative neurons, and the rate maps less clear, for which one might surmise that these neurons need not necessarily belong to the module originally assigned to.

\section{Additional visualizations for Head Direction and Grid Cell Data}\label{section:additional_visualizations_real_data}

\begin{figure}
\floatbox[{\capbeside\thisfloatsetup{capbesideposition={right,center},capbesidewidth=0.36\textwidth}}]{figure}[\FBwidth]
{\caption{\textbf{Supplementary visualizations for analyses on real datasets}. \textbf{A}, Recorded mouse head direction ('True HD') and inferred latent trajectories on test data. Results from the experiment on head direction data, Sec. \ref{sec:head_cells}. \textbf{B}, Dynamics of the inferred toroidal coordinates plotted as a function of the $(x, y)$ spatial position of the animal (two tori, each with two circular coordinates, hence four figures). Results from the experiment on grid cell data, Sec. \ref{sec:grid cells data}.}\label{More_real_data_visualization_stuff}}
{\includegraphics[width=0.6\textwidth]{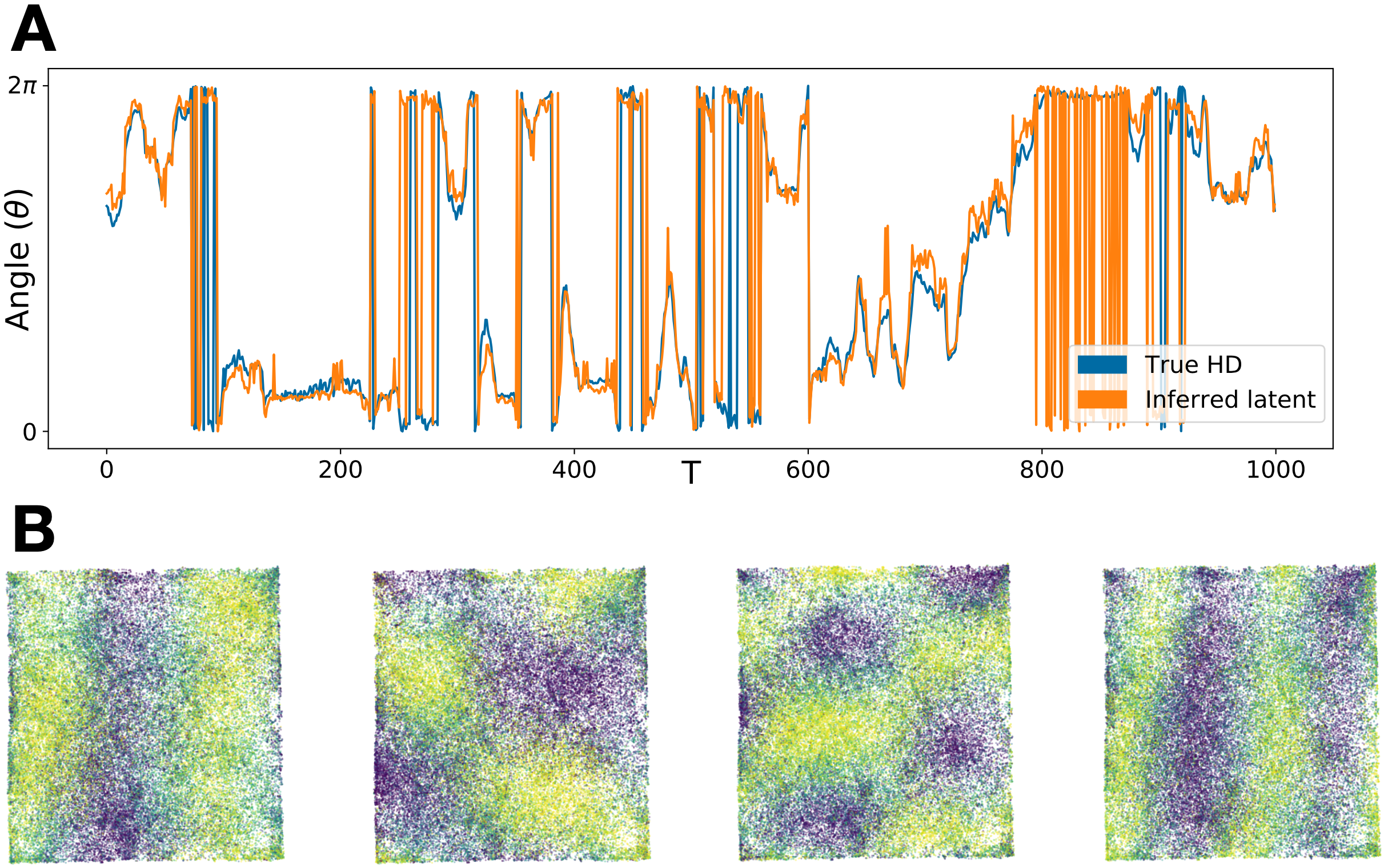}}
\end{figure}

To supplement the results shown in Sec. \ref{sec:head_cells} and Sec. \ref{sec:grid cells data}, we have included additional figures for both the head direction application and the grid cell application. More precisely, Fig. \ref{More_real_data_visualization_stuff}A shows the inferred variational mean of faeLVM-n plotted against the recorded mouse head direction (now as trajectories, in contrast to the scatterplot in Fig. \ref{Application_to_peyrache}A, while Fig. \ref{More_real_data_visualization_stuff}B shows inferred latent dynamics on grid cell data, as function of spatial position. As there is no way of recording the 'true' toroidal coordinates (akin to what is done in the head direction dataset), this figure is included in an effort to demonstrate the clear correspondence between each of the inferred toroidal axes and the spatial position, with the results being in agreement with what was found in \citet{gardner2022toroidal}.

\section{Extended ensemble detection simulation}\label{section:additional_ensembles}

In this section we present an additional experiment related to the ensemble detection task. Namely, we ask the model to separate five distinct ring-like ensembles, $z_j \in \mathbb{S}^1$, to complement the experiments from Section \ref{sec:ensemble_detection}, which only considered two and three ensembles. 

\begin{figure}
  \centering
  \includegraphics[width=1\textwidth]{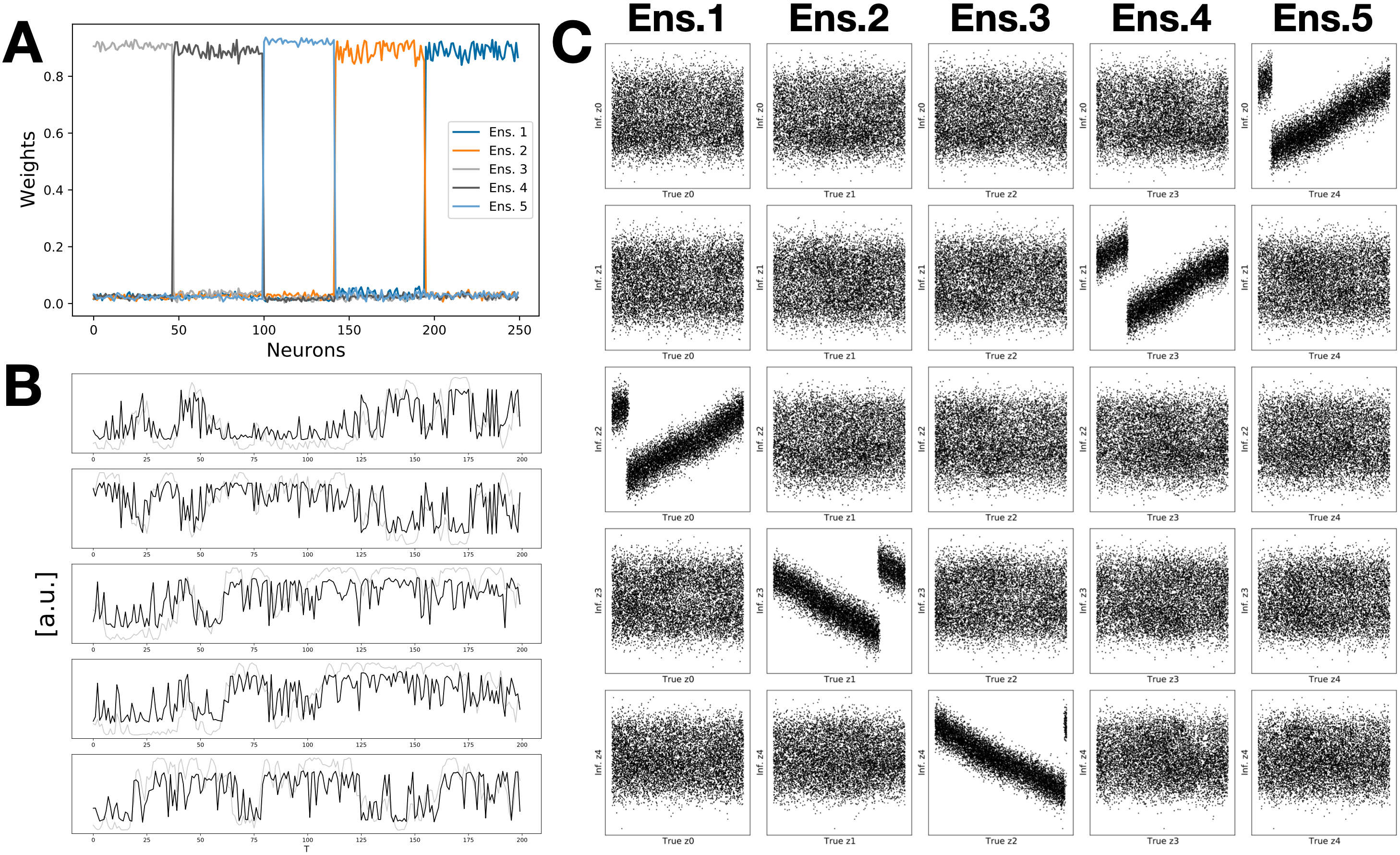}
  \caption{\textbf{Ensemble detection of 5 circles}. \textbf{A}, Learned ensemble weights for the neurons belonging to five unique circular ensembles. \textbf{B}, True spike rate (grey) vs predicted spike rate (black) for for five randomly selected test neurons, one from each ensemble. \textbf{C}, Scatter plot of true latent trajectory against inferred latent trajectory.}
  \label{Extra ensemble detection}
\vspace{10pt}
\end{figure}

From Fig. \ref{Extra ensemble detection}, we can see that faeLVM-b is able to separate the five ensembles quite clearly. The weights in Fig. \ref{Extra ensemble detection}A is approaching the desired one-hot encoding, and we see that the five inferred ensembles correspond very nicely with the true ensembles (Fig. \ref{Extra ensemble detection}C). The model is also able to accurately predict spike rates on held out neurons from each ensemble (\ref{Extra ensemble detection}B). Thus, this simulation provides evidence that our model is also able to work on more challenging ensemble separation tasks, and is not only limited to lower ensemble numbers.


\section{Hybrid inference}
\label{section: hybrid_inference}

\subsection{Overview}

Here, we give an overview on how the hybrid inference is performed. In contrast to the traditional variational approach, where one uses a single sample from the variational posterior as prediction for the latent variable, we propose drawing a selection of samples, then performing gradient descent on said latents using the Adam optimizer.
Specifically, if $z$ are the latents for the test set (as inferred by the encoder and variational posterior) and $x$ the responses of the training neurons on the test set, we maximize
\begin{equation}
    z^* = \underset{z}{\max} \log p(x|z)
\end{equation}
for $2000$ steps with the Adam optimizer and a learning rate of $0.001$.
We perform this in parallel on $M=10$ distinct samples from the variational posterior and pick the one with the highest final likelihood (note that all samples usually converge to the same value).
We emphasize that all model parameters are fixed during hybrid inference, and the optimization only affects the sampled latents.

Empirically, we observe that this not only improves the likelihood for the training neurons (i.e., the objective we are maximizing), but crucially also the likelihood of the test neurons.
This means that we are not just overfitting but actually improving our point estimates of the latents in this inference procedure.
We leave optimization of the full likelihood, including uncertainty estimates on the latents, for future research.

\subsection{Hybrid Inference simulations}

\begin{table}
  \caption{\textbf{Hybrid inference comparison}. Table showing mean negative log-likelihood (NLLH) and mean geodesic error (GE) for the six models specified (lower is better for both criterions), under varying amounts of synthetically generated training data with different numbers of training neurons ($N$) and different training dataset lengths ($T$).}
  \label{VAE_table}
  \centering
  \begin{tabular}{lllllllllll}
    \toprule
    \multicolumn{1}{r}{$N$ / $T$} &
    \multicolumn{2}{c}{$30$ / $100$} &
    \multicolumn{2}{c}{$30$ / $500$} &
    \multicolumn{2}{c}{$30$ / $1000$} &
    \multicolumn{2}{c}{$15$ / $1000$} &
    \multicolumn{2}{c}{$45$ / $1000$}\\
    \cmidrule(r){2-3}
    \cmidrule(r){4-5}
    \cmidrule(r){6-7}
    \cmidrule(r){8-9}
    \cmidrule(r){10-11}
        LVM & NLLH & GE & NLLH & GE & NLLH & GE & NLLH & GE & NLLH & GE\\
    \midrule
    v-fae-n & $12745$  & $0.57$ & $8971$ & $0.28$ & $8781$ & $0.27$ & $9110$ & $0.45$ & $8642$ & $0.27$ \\
    v-fae-s & $11798$  & $0.64$ & $9167$ & $0.34$ & $8949$ & $0.31$ & $9718$ & $0.40$ & $8820$ & $0.31$ \\
    v-fae-b & $10912$ & $0.50$ & $9152$ & $0.36$ & $8830$ & $0.30$ & $9492$ & $0.33$ & $8633$ & $0.26$ \\
    \midrule
    fae-n & $11975$  & $0.58$ & $8775$ & $0.20$ & $8608$ & $0.17$ & $9018$ & $0.44$ & $8415$ & $0.09$ \\
    fae-s & $10963$ & $0.42$ & $8694$ & $0.13$ & $8657$ & $0.14$ & $9460$ & $0.29$ & $8491$ & $0.13$ \\
    fae-b & $10040$ & $0.25$ & $8636$ & $0.12$ & $8501$ & $0.12$ & $9227$ & $0.23$ & $8267$ & $0.05$ \\
    \bottomrule
  \end{tabular}
\end{table}

For evaluate of the performance gain with hybrid inference at test-time, we consider six variations of the faeLVM; that being the three cases described in Sec. \ref{sec:methods}, each alternative including a case of using hybrid inference and one using the standard variational posterior approach (fae vs v-fae).

As in Sec. \ref{sec:feature_sharing_sims}, we divide our data into four parts (Fig. \ref{Train_Test}), generate a $1$D periodic latent variable and simulate spikes according to a Poisson distribution with Gaussian tuning curves. We fix $T_{\text{test}}$ and $N_{\text{test}}$ ($1000$ and $30$), vary $T_{\text{train}}$ and $N_{\text{train}}$ over a selection of values, and for each condition, repeat the experiment over $20$ seeds (synthetic data is different for each seed, but all of the models are trained on the same dataset, to ensure that comparisons are still valid), reporting mean negative log-likelihood (NLLH) and mean geodesic error (GE) on test data. 

The results presented in Table \ref{VAE_table} indicate that performance, in almost every case, increases substantially by including the hybrid inference step. Hence, all other experiments performed in this article are done solely using the hybrid inference where applicable (e.g., when applying the model to grid cell data, evaluations are done in-training-sample, and without the focus on latent prediction on a test set, hence there is no need for hybrid inference at test-time).

\section{Hyperparameter search}\label{section: hyperparams}

To investigate hyperparameters and their effect, we selected relevant parameters and estimated their ranges based on related literature. The relevant parameters can be seen on the left in Table~\ref{tab:hyperparameter}. 

To determine optimal values of hyperparameters, we relied on a random search strategy. A grid search is not feasible due to the curse of dimensionality, and a random search can also be more effective \citep{bergstra2012random}. Furthermore, the random search can easily be used for discrete and continuous variables, that could be on linear or logarithmic scales. The selected scales and ranges, can be found in the middle in Table~\ref{tab:hyperparameter}. 

In total, we sample $500$ hyperparameter configurations. Next, we train and test each configuration for 3 random seeds. We evaluate the models on the test accuracy and mean correlation. The final selected model is shown on the far right in Table~\ref{tab:hyperparameter}. 

\begin{figure}[h]
  \centering
  \includegraphics[width=1\textwidth]{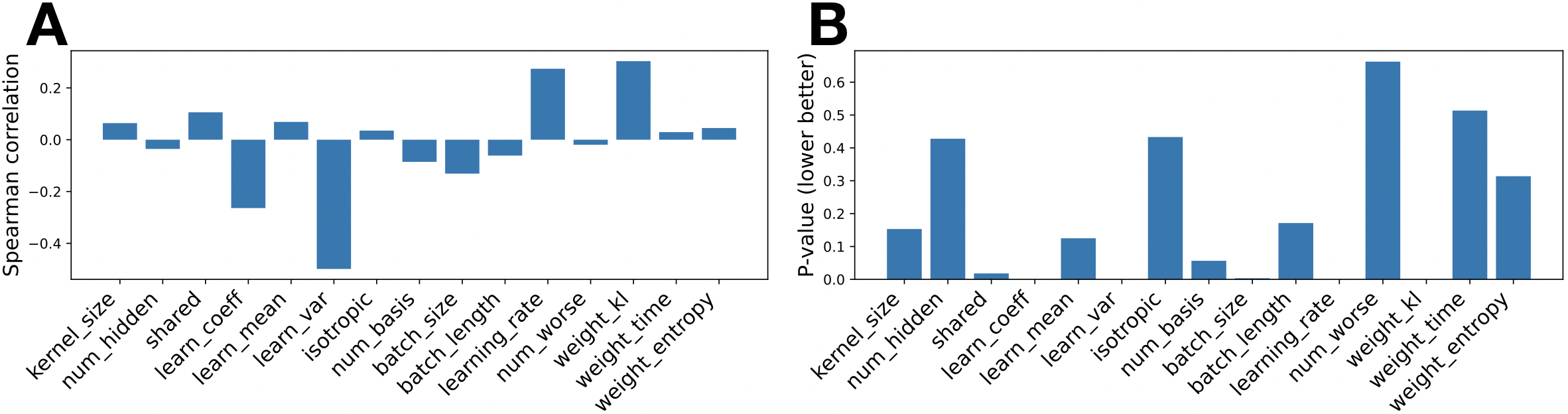}
  \caption{\textbf{Correlations with test LLH}. \textbf{A}, Spearman correlation. \textbf{B}, P-values.}
  \label{fig:hyper_correlations}
\end{figure}

Lastly, we also measure whether there is an ordinal correlation (Spearman) between the varied factors and the test LLH. The results are shown in Fig.~\ref{fig:hyper_correlations}. We see that learning the variance (learn\_var) increases the performance on average. Also, lower learning rates and larger batch-sizes lead to better results. Lastly, also lower weighting for the KL term leads to better results. 
\newpage
\begin{table}[h]
    \begin{tabular}{lll}
    \toprule
    Hyperparameter &                              Ranges &     Best \\
    \midrule
       kernel\_size &                  choice([1, 9, 17]) &         1 \\
        num\_hidden & choice([16, 32, 64, 128, 256, 512]) &       128 \\
            shared &               choice([True, False]) &     False \\
       learn\_coeff &               choice([True, False]) &      True \\
        learn\_mean &               choice([True, False]) &     False \\
         learn\_var &               choice([True, False]) &     False \\
         isotropic &               choice([True, False]) &     False \\
         num\_basis &                choice([1, 2, 4, 8]) &         4 \\
      nonlinearity &             choice([exp, softplus]) &  softplus \\
        batch\_size &                 choice([1, 16, 32]) &        32 \\
      batch\_length &                   choice([64, 128]) &        64 \\
     learning\_rate &              loguniform(1e-4, 1e-1) &  0.001003 \\
         num\_worse &               choice([10, 50, 100]) &        10 \\
         weight\_kl & choice([0., loguniform(1e-9, 1e0)]) &       0.0 \\
       weight\_time & choice([0., loguniform(1e-9, 1e0)]) &       0.0 \\
    weight\_entropy & choice([0., loguniform(1e-9, 1e0)]) &       0.0 \\
    \bottomrule
    \end{tabular}
    \caption{\textbf{Hyperparameter search.} We show our selected hyperparameters and their corresponding ranges.}
    \label{tab:hyperparameter}
\end{table}

\end{document}